# Particle Swarm Optimization with State-Based Adaptive Velocity Limit Strategy

Xinze Li[1], Kezhi Mao[1], Fanfan Lin[2], Xin Zhang[3]

[1]*School of Electrical and Electronic Engineering, Nanyang Technological University, Singapore 639798, Singapore*
[2]*Interdisciplinary Graduate School, Nanyang Technological University, Singapore 639798, Singapore*
[3]*College of Electrical Engineering, Zhejiang University, Hangzhou 310027, China*
**Corresponding Author: Kezhi Mao (Email: EKZMao@ntu.edu.sg)**

**Abstract**

Velocity limit (VL) has been widely adopted in many variants of particle swarm optimization (PSO) to prevent particles from searching outside the solution space. Several adaptive VL strategies have been introduced with which the performance of PSO can be improved. However, the existing adaptive VL strategies simply adjust their VL based on iterations, leading to unsatisfactory optimization results because of the incompatibility between VL and the current searching state of particles. To deal with this problem, a novel PSO variant with state-based adaptive velocity limit strategy (PSO-SAVL) is proposed. In the proposed PSO-SAVL, VL is adaptively adjusted based on the evolutionary state estimation (ESE) in which a high value of VL is set for global searching state and a low value of VL is set for local searching state. Besides that, limit handling strategies have been modified and adopted to improve the capability of avoiding local optima. The good performance of PSO-SAVL has been experimentally validated on a wide range of benchmark functions with 50 dimensions. The satisfactory scalability of PSO-SAVL in high-dimension and large-scale problems is also verified. Besides, the merits of the strategies in PSO-SAVL are verified in experiments. Sensitivity analysis for the relevant hyper-parameters in state-based adaptive VL strategy is conducted, and insights in how to select these hyper-parameters are also discussed.

## 1. Introduction

Particle swarm optimization (PSO), firstly proposed in [1], is one of the population-based algorithms. Inspired by the natural swarming behavior of bird flocks, PSO particles will be guided towards the optimal solution by the iterative update of velocity and position [2]. In comparison to other popular population-based algorithms, such as genetic algorithm [3] and ant colony optimization [4], PSO enjoys faster convergence speed, since it utilizes the global best information directly in guiding the searching process. PSO is perfectly appropriate for continuous optimization [5], and is widely used in applications such as the selection of neural network structure [6–8], scheduling problems [9][10], and dynamic optimization problems [11][12][13].

Generally, the variants of PSO can be divided into three types. The first type is the PSO variants with adaptive adjustments of hyper-parameters. For instance, the hierarchical PSO integrated with time-varying acceleration coefficients (HPSO-TVAC) is proposed by Ratnaweera et al. [14] and adaptive PSO (APSO) by Zhan et al. [15] adaptively adjusts the acceleration coefficients during searching. Adaptive granularity learning distributed PSO (AGLD-PSO) proposed by Wang et al. [16] and multi-swarm PSO (MPSO) proposed by Xia et al. [17] separate the original swarm into several small sub-swarms and adaptively tune the size of sub-swarms. In the fitness-based differential evolution (FADE) proposed by Xia et al. [18], the population size is modified to be adaptive.

The second type is the variants with different learning strategies, which are increasingly important and popular in PSO variants. The equation of velocity update has been modified by Liang et al. in the comprehensive learning PSO (CLPSO) [19] and by Liu et al. in randomly occurring distributed delayed PSO (RODD-PSO) [20]. Also, particles' velocities are updated with three exemplar archives by Xia et al. in triple archives PSO (TAPSO) [21]. And Li et al. adopted parallel update for particles in pipeline-based parallel PSO (PPPSO) [22]. Apart from these, the popular orthogonal learning strategy is also an example of the second type, which enables the swarm to fly towards better directions and increases the chance of finding global optimum. The algorithms related with this strategy include the orthogonal learning PSO (OLPSO) proposed by Zhan et al. [23], orthogonal learning brain storm optimization (OLBSO) proposed by Ma et al. [24] and orthogonal local search genetic algorithm (OLSGA) proposed by Huang et al. [25].

In the third type, apart from the original steps in PSO, extra auxiliary strategies such as limit handling strategy [26][27], restart strategy [28] and Gaussian mutation strategies [29] have been put forward. For example, cooperative coevolutionary bare-bones PSO (CCBBPSO) proposed by Zhang et al. [30] utilizes decomposition methods to reduce the complexity of optimization problems. Besides, bottleneck objective learning strategy and elitist learning and juncture learning strategies are provided in coevolutionary PSO (CPSO) by Liu et al. [31]. Additionally, dynamic group learning strategy is another typical example which is introduced by Wang et al. in dynamic group learning distributed PSO (DGLD-PSO) [32].

In many PSO variants, the velocities of particles are clamped by velocity limit (VL), which prevents particles from searching outside the solution space [33][34]. Initially, only fixed VL is considered during the iterations of PSO [35]. Recently, adaptive VL strategies have been introduced in some PSO variants since the performance of PSO can be improved if VL is properly tuned during the iterations. For instance, in [36], VL can linearly, sectionally and exponentially decrease as the number of iterations increases. In

[37], polynomial decreasing of VL is illustrated. In [38], VL is updated based on particles' outermost position. Generally, the existing adaptive VL strategies in PSO variants suffer from nontrivial problems which is attributable to that VL is adjusted based on the number of iterations, detailly discussed as follows.

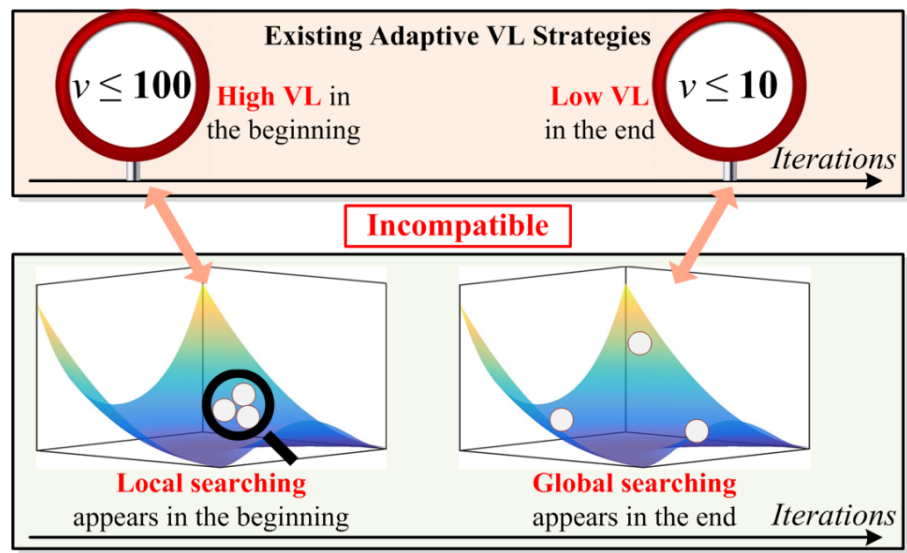

**Fig. 1.** Intuitive explanation of the incompatibility between the existing adaptive VL strategies and the current searching state of particles.

The problem of the existing adaptive VL strategies is the incompatibility between these adaptive VL strategies and the current searching state of particles, resulting in unsatisfactory optimization results, as shown in Fig. 1. As stated in [39], a low VL is beneficial for local searching, and a high VL is helpful for global searching. And the value of VL decreases during the iterations in most of the existing adaptive VL strategies [36][37]. Thus, these strategies are beneficial for global searching in initial iterations with a high value of VL, and they are helpful for local searching in final iterations with a low value of VL. However, the current state of particles is volatile and non-deterministic [15], indicating that local searching can occur at the beginning of the iterations, and global searching can occur in the final iterations. Hence, the existing adaptive VL strategies can be incompatible with the current state of particles. Due to the incompatibility, the performance of the PSO variants with these strategies will be undermined. As a result, **how to adaptively tune VL to match the current searching state of particles is the main challenge to be considered in this paper**.

Aimed at the considered challenge, to match the tuning of VL with the current searching state of particles during the iterations, a novel PSO variant with state-based adaptive VL strategy (PSO-SAVL) is proposed, in which the evolutionary state estimation (ESE) introduced by Zhan et al. in APSO [15] is adopted to indicate the current state of particles. In the strategy of ESE, the positional distribution and fitness values of particles are considered to decide which state the swarm belongs to currently. Besides, in the proposed PSO-SAVL, to improve the capability of avoiding local optima, limit handling strategies

[27] are adopted and are combined with ESE, which adjust the positions and velocities of particles inside their limits when the limits are exceeded.

The organization of this paper is as follows. In Section 2, the functions of VL and ESE are introduced. In Section 3, the major challenge in the existing adaptive VL strategies is elaborated. In Section 4, the process of the proposed PSO-SAVL\is discussed, then the state-based adaptive VL strategy and the limit handling strategies are illustrated, and the complexity of PSO-SAVL is analyzed. In Section 5, experimental results of the proposed PSO-SAVL compared with other popular PSO variants on a wide range of standard benchmark functions of 50 dimensions are detailly analyzed, and its scalability on high-dimension and large-scale problems are also studied. In Section 6, the merits of the state-based adaptive VL and limit handling strategies in PSO-SAVL are studied, the sensitivity of newly introduced hyper-parameters in the proposed PSO-SAVL is thrown light on, and insights in how to select these hyper-parameters are also discussed. Finally, conclusions are summarized in Section 7.

## 2. Preliminaries of Velocity Limit and Evolutionary State Estimation

In this section, the functions of VL, and ESE which can reflect the current searching state of particles are introduced.

### 2.1 Preliminaries: Functions of Velocity Limit

In PSO variants, VL is usually adopted to clamp the velocities of particles. The functions of VL can be summarized as: to prevent PSO from searching outside the solution space, to accelerate the convergence speed, and to adjust the tradeoff between local searching and global searching. In detail, compared with the variants without VL, the variants with the clamping of VL can restrict the magnitude of velocity, prevent the particles from flying far away, and thus prevent them from searching outside the solution space [40]. With the clamping of VL, convergence speed is faster [41]. Moreover, since a high VL is helpful for global searching and a low VL is beneficial for local searching, VL can adjust the tradeoff between local searching and global searching [39]. An approximated theoretical analysis is given as follows.

As analyzed by Maurice and James [42], the velocity in the reduced dynamic equation of PSO without VL is shown in (1):

$$v_{t+1} = v_t + \varphi(p - x_t), \tag{1}$$

where the problem space is one-dimension, the population includes one particle, $p$ and $\varphi$ are constant, and $v_t$ and $x_t$ are the velocity and the position of the particles in the $t^{th}$ iteration.

Since VL restricts $v_t$ whenever $v_t$ exceeds its limit, the effects of VL can be represented by an equivalent discount factor $a$, the range of which lies within [0, 1]. Thus, the velocity in the reduced dynamic equation of PSO with VL can be described in (2), where the discount factor $a$ is regarded as constant for analytical convenience.

$$v_{t+1} = a\left(v_t + \varphi(p - x_t)\right), a \in [0,1]. \qquad (2)$$

Due to the accumulated discount effects of $a$ in each iteration, the magnitude of $v_{t+1}$ of the variants with VL is much smaller than the magnitude of $v_{t+1}$ of the variants without VL, and thus the change of position in the variants with VL will be smaller, leading to faster convergence speed.

Furthermore, when compared with a high value of VL, a lower value of VL is equivalent to a smaller $a$, resulting in the smaller magnitude of $v_{t+1}$ and the smaller change of position, which encourages PSO to search for the local space thoroughly. In a word, a lower value of VL can benefit local searching.

Similarly, in comparison to a low value of VL, a higher value of VL can be represented by a larger $a$, because of which the magnitude of $v_{t+1}$ is larger, and the change of position is larger, leading to the wider exploration of solution space. The wider exploration of solution space encourages PSO to globally search for the best solution. Hence, a high VL is beneficial for global searching.

In a word, if VL is adaptively tuned based on the current state of particles to match global searching and local searching during the iterations, the performance of PSO can be improved.

## 2.2 Preliminaries: Introduction to Evolutionary State Estimation and its Relationships with VL

ESE technique has been firstly provided in APSO [15] by Zhan et al., in which the population distribution and fitness values of particles are adopted to evaluate the evolutionary state of the swarm. According to Zhan et al., the mean distance from the globally best particle to other particles ($d_g$) is relatively larger in exploration states and smaller in exploitation states. To scale $d_g$ to a range of [0,1], equation (4) is applied to $d_g$ to define evolutionary factor $f$. $f$ is utilized to indicate the four states of swarm during evolutionary process: exploration, exploitation, convergence and jumping-out. In a word, the classification of the state of particles depends on the value of evolutionary factor $f$, which is computed as the following:

$$d_i = \frac{1}{N-1} \sum_{j=1, j \neq i}^{N} \sqrt{\sum_{d=1}^{D} \left(X_i^d - X_j^d\right)^2}, \qquad (3)$$

$$f = \frac{d_g - d_{min}}{d_{max} - d_{min}}, \qquad (4)$$

where the position of the $i^{th}$ particle is represented by a vector $X_i = [X_i^1, X_i^2, \ldots, X_i^D]$, $D$ stands for the dimension of the solution space, $N$ is population size, $d_i$ is the mean distance from the $i^{th}$ particle to other particles, $d_g$ is mean distance from the globally best particle to other particles, and $d_{min}$ and $d_{max}$ are the minimum and maximum values of all $d_i$. Additionally, ESE strategy has been recently further improved in ADDE [43] by Zhan et al. which significantly decreases its computational burden.

**Table 1.**

Searching State of Particles.

| | State | $f$ |
|---|---|---|
| Local Searching | Convergence | [0, 0.25) |
| | Exploitation | [0.25, 0.5) |
| Global Searching | Exploration | [0.5, 0.75) |
| | Jumping-out | [0.75, 1] |

It is of great significance to divide the evolutionary states of swarm to these four types because different evolutionary states may indicate different searching states, as listed in Table 1. If the swarm is in convergence and exploitation states, they are experiencing the local searching process, in which more preciseness is demanded. In this case, it is reasonable to assume that low-value VL will be helpful for accuracy in local searching. It is because that a small VL is able to realize a relatively small change in every position update which is able to realize higher preciseness in searching. While if the swarm is in exploration and jumping-out states, they are in the process of global searching, in which wider searching range is expected. On this condition, it is understandable that high-value VL can benefit a stronger global searching capability for that a relatively big change in every position update contributes to lower risk of neglecting global optima.

However, according to the analysis in [15], it should be noted that the state of swarm is non-deterministic during the iterations of PSO, which means local and global searching can occur in any iteration and their appearance does not follow a fixed sequence. This uncertainty of the evolutionary state of the swarm brings challenges for the tuning of VL.

## 3. Challenge in the Existing Adaptive Velocity Limit Strategies

In the existing adaptive VL strategies, there is one major problem: these strategies are incompatible with the current searching state of swarm. To elaborate the incompatibility problem, two existing adaptive VL strategies are given in Fig. 2, as well as the evolutionary factor $f$ of PSO-LDIW [44] under the Rosenbrock's function.

According to the elaboration of the functions of VL in Section 2, a high VL is beneficial for global searching and local searching favors a low VL. The two strategies in Fig. 2 decreases their VL during evolutionary process based on some functions of iteration number, which leads to a result that VL is updated in a deterministic way. These strategies are based on the assumption that the appearance of global search and local search will follow a deterministic sequence. Thus, these strategies are beneficial for global searching in initial iterations, and are helpful for local searching in final iterations [39].

However, the searching state of particles is non-deterministic, which means that local searching may occur in initial iterations and global searching may occur in final iterations. As shown in Fig. 2, the evolutionary factor $f$ in several initial iterations is lower than 0.5, indicating that local searching states appear in the beginning, and $f$ in final iterations is higher than 0.5, indicating that global searching occur in last iterations. Consequently, as can be seen from Fig. 2, the local searching states appeared in initial iterations are incompatible with the assigned high VL, and the global searching states occurred in final iterations are incompatible with the assigned low VL. The incompatibility undermines the performance of the PSO with the existing adaptive VL strategies.

The incompatibility problem between the existing adaptive VL strategies and the current state of particles as discussed above can always occur, leading to unsatisfactory performance of PSO. **As a result, the major challenge in the existing adaptive VL strategies is how to adaptively tune VL to match the current searching state of particles.**

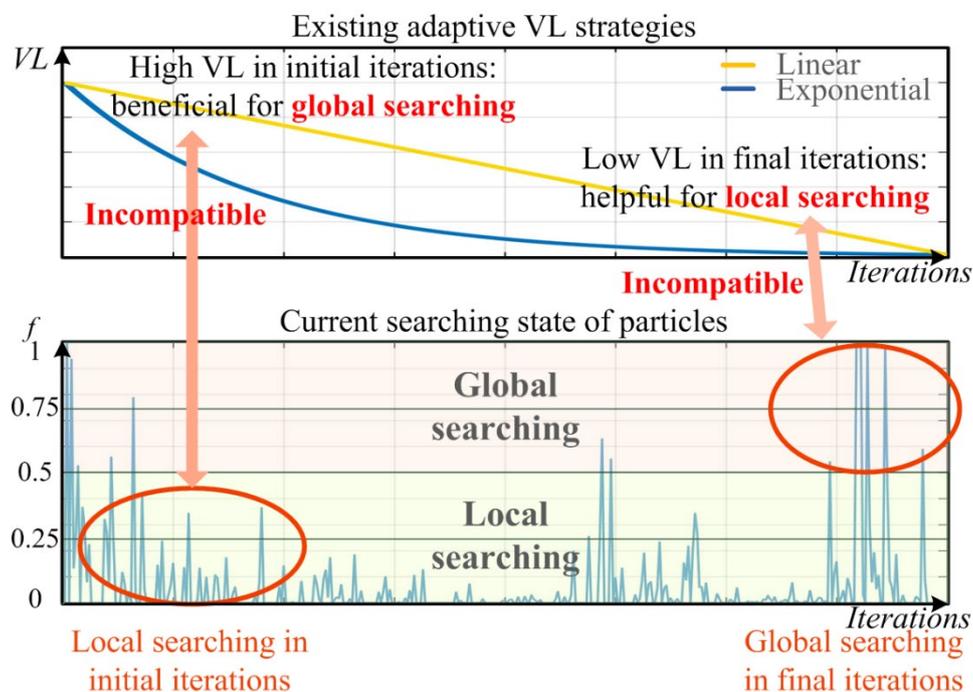

**Fig. 2.** Problems of the existing adaptive VL strategies: incompatibility between the setting of VL and the current searching state of particles.

## 4. The Proposed Novel PSO Variant: PSO-SAVL

In this section, a novel PSO variant with state-based adaptive velocity limit strategy is proposed. In the proposed PSO-SAVL, to solve the challenge in the existing adaptive VL strategies, a state-based adaptive VL (SAVL) strategy is proposed. Apart from that, position and velocity limit handling strategies are adopted and modified to improve the capability of avoiding local optima.

## 4.1 Process of the Proposed PSO-SAVL

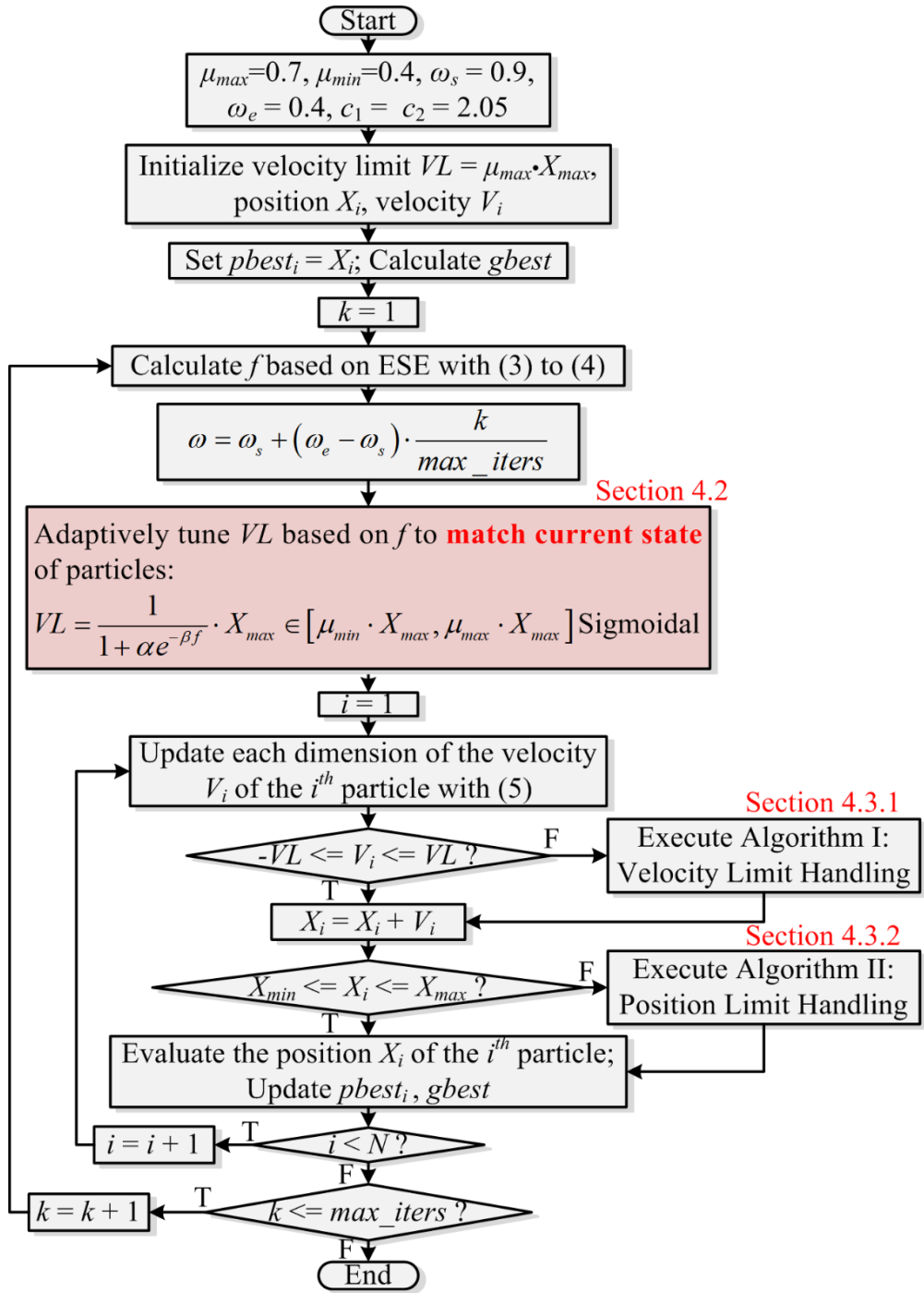

**Fig. 3.** Flowchart of the proposed PSO-SAVL.

The flowchart of PSO-SAVL is shown in Fig. 3, in which the velocity of the $d^{th}$ dimension of the $i^{th}$ particle $V_i^d$ is updated with:

$$V_i^d = \omega \cdot V_i^d + c_1 \cdot r_{1,i}^d \cdot \left( pbest_i^d - X_i^d \right) + c_2 \cdot r_{2,i}^d \cdot \left( gbest^d - X_i^d \right), \tag{5}$$

where $r_{1,i}^d$ and $r_{2,i}^d$ are two uniformly distributed random numbers within [0, 1] of the $d^{th}$ dimension of the $i^{th}$ particle. The position and velocity of the $i^{th}$ particle are represented by vectors $X_i = [X_i^1, X_i^2, \ldots,$

$X_i^D$] and $V_i = [V_i^1, V_i^2, …, V_i^D]$. $D$ is the dimension of the solution space. The limits of position and velocity are represented by vectors $X_{max} = [X_{max}^1, X_{max}^2, …, X_{max}^D]$, $X_{min} = [X_{min}^1, X_{min}^2, …, X_{min}^D]$ and $VL = [VL^1, VL^2, …, VL^D]$. $\mu_{max}$ and $\mu_{min}$ are the newly-introduced hyper-parameters, representing the maximum and minimum proportions of $VL$ to $X_{max}$, respectively, and thus $VL$ lies within $[\mu_{min} \cdot X_{max}, \mu_{max} \cdot X_{max}]$. $pbest_i$ and $gbest$ are the historical best position of the $i^{th}$ particle and the global best position, respectively. $\omega$ is the inertia weight. $c_1$ and $c_2$ are the acceleration coefficients. $max\_iters$ is the maximum number of iterations. $N$ is the population size.

In Fig. 3, the red-highlighted block shows the proposed state-based adaptive VL strategy, which adjusts VL following the sigmoidal function of $f$ to match the current state of particles. Overall, in the $k^{th}$ iteration, the current state of particles is reflected by $f$ based on ESE with (3) to (4), and then, VL is adjusted based on $f$. After the adjustment of VL, the velocities of particles can be updated and then clamped by the newly-adjusted VL with Velocity Limit Handling algorithm. Afterwards, the position is updated and clamped with Position Limit Handling algorithm. The iteration of PSO-SAVL continues until $k$ reaches $max\_iters$.

## 4.2 The State-Based Adaptive VL Strategy

### 4.2.1 Adaptive Tuning of VL According to the Sigmoidal Function of Evolutionary Factor $f$

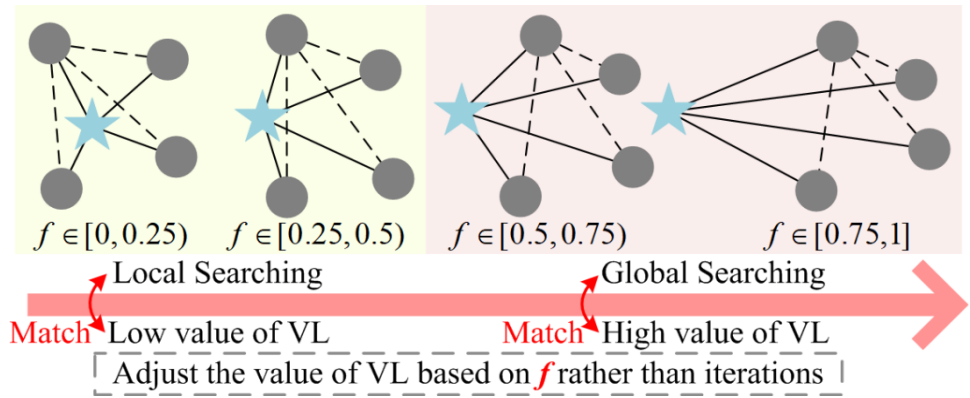

**Fig. 4.** The relations between the value of VL and the evolutionary factor $f$, blue star is the position of globally best particle, and grey circle is the position of other particles.

As stated in Section 3, the incompatibility between the existing adaptive VL strategies and the current state of particles will lead to worse performance of PSO variants than the variants with an adaptive VL strategy that can match the current state of particles. Hence, it is of significance to adjust $VL$ to match the current searching state of particles.

To realize adaptive $VL$, $\mu$ is defined as the proportion of velocity limit ($VL$) to the maximum of particles' position ($X_{max}$) as shown in (6). $VL$ will be always maintained within the range [$VL_{min}$, $VL_{max}$], which can also be expressed with the minimum proportion $\mu_{min}$ and maximum proportion $\mu_{max}$, as described with (7).

$$VL = \mu \cdot X_{max} \tag{6}$$

$$VL \in [VL_{min}, VL_{max}] = [\mu_{min} \cdot X_{max}, \mu_{max} \cdot X_{max}] \tag{7}$$

As shown in Fig. 4, since the current searching state of particles can be reflected by the evolutionary factor $f$ in (4), $VL$ should be adjusted based on $f$, rather than iterations. To make $VL$ adapt to $f$, $\mu$ is chosen to be the sigmoidal mapping function of $f$, and $\alpha$ and $\beta$ are the hyper-parameters in $sigmoid(f)$, as shown in (8). From evolutionary factor $f$ point of view, when $f = 0$ and $f = 1$, $VL$ reaches its minimum $VL_{min}$ and its maximum $VL_{max}$, respectively, as described in (9).

$$VL = \mu \cdot X_{max} = sigmoid(f) \cdot X_{max} = \frac{1}{1 + \alpha e^{-\beta f}} \cdot X_{max} \tag{8}$$

$$VL_{min} = sigmoid(f = 0) \cdot X_{max} = \frac{1}{1 + \alpha} \cdot X_{max} \tag{9a}$$

$$VL_{max} = sigmoid(f = 1) \cdot X_{max} = \frac{1}{1 + \alpha e^{-\beta}} \cdot X_{max} \tag{9b}$$

Thus, on the basis of (7) and (9), if $\mu_{min}$ and $\mu_{max}$ have been set, the relationships between $\alpha$ and $\mu_{min}$, as well as the relationships between $\beta$ and $\mu_{max}$ will be obtained, as shown in (10a) and (10b).

$$VL_{min} = \mu_{min} \cdot X_{max} = \frac{1}{1 + \alpha} \cdot X_{max} \Rightarrow \mu_{min} = \frac{1}{1 + \alpha} \tag{10a}$$

$$VL_{max} = \mu_{max} \cdot X_{max} = \frac{1}{1 + \alpha e^{-\beta}} \cdot X_{max} \Rightarrow \mu_{max} = \frac{1}{1 + \alpha e^{-\beta}} \tag{10b}$$

Thus, the hyper-parameters $\alpha$ and $\beta$ can be calculated with (11) after the setting of $\mu_{min}$ and $\mu_{max}$. And the setting of $\mu_{min}$ and $\mu_{max}$ will be detailly described in Section 6.2.

$$\alpha = \frac{1}{\mu_{min}} - 1 \tag{11a}$$

$$\beta = -\ln\left(\left(\frac{1}{\mu_{max}} - 1\right) / \alpha\right) \tag{11b}$$

The evolutionary factor $f$ and $VL^1$ of PSO-SAVL on 10-D Rosenbrock's function are plotted in Fig. 5, in which the $\mu_{min}$ and $\mu_{max}$ are selected as 0.4 and 0.7 respectively. Obviously, the value of VL matches well with the evolutionary factor $f$ during the iterations. Hence, with the proposed state-based adaptive VL strategy in (8), the changing of VL can match the current state of particles, and the incompatibility between the existing adaptive VL strategies and the current searching state of particles is solved.

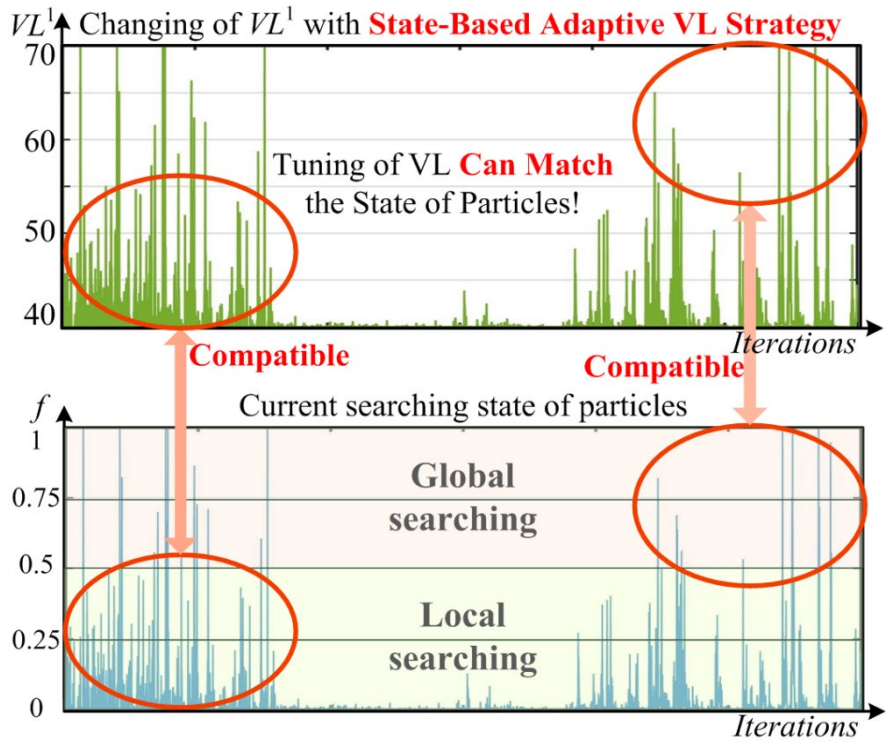

**Fig. 5.** $f$ and $VL^1$ of PSO-SAVL on 10-D Rosenbrock's function: (a) $f$; (b) $VL^1$.

### 4.2.2 Advantages of the Proposed State-Based Adaptive VL Strategy

**Table 2.**

Test Functions Used in this Paper.

| Function | Position Limits | Global $f_{min}$ | Acceptance |
|---|---|---|---|
| Sphere $f_1$ | $[-100^D, 100^D]$ | 0 | 0.01 |
| Rosenbrock $f_2$ | $[-100^D, 100^D]$ | 0 | 500 |
| Rastrigin $f_3$ | $[-5.12^D, 5.12^D]$ | 0 | 50 |
| Griewank $f_4$ | $[-600^D, 600^D]$ | 0 | 0.5 |
| Schwefel $f_5$ | $[-500^D, 500^D]$ | 0 | 7000 |
| Rotated Griewank $f_6$ | $[-600^D, 600^D]$ | 0 | 5 |
| Rotated Rastrigin $f_7$ | $[-5.12^D, 5.12^D]$ | 0 | 150 |

Generally, the PSO variants with the proposed state-based adaptive VL strategy in (8) can achieve better optimization solutions and faster convergence speed than the PSO variants with existing adaptive VL strategies. These advantages are experimentally verified as the followings.

In this paper, to benchmark the performance of PSO variants, 7 representative functions [13, 26, 32] are adopted, which cover a wide range of types. As summarized in Table 2, $f_1$ is unimodal function, $f_2$ to $f_5$ are unrotated multimodal functions, and $f_6$ to $f_7$ are complex rotated multimodal functions. All functions are required to be minimized, and their global minimum $f_{min}$ is shown in Table 2, where the "Acceptance" is the threshold to judge whether a solution is acceptable or not.

To test the advantages of the proposed state-based adaptive VL strategy, the fixed VL strategy and an existing iteration-dependent adaptive VL strategy (the linearly decreasing VL) [36] are compared with. PSO-LDIW [44], FDR-PSO [46], APSO [15], and RODD-PSO [20], adopting one of the fixed VL, the iteration-dependent VL and the proposed state-based adaptive VL strategies, run on the test functions $f_2$, $f_3$, $f_5$ and $f_6$ over 30 independent trials. In this part, dimension $D$ is 10, maximum iterations $max\_iters$ is 3000, population number $N$ is 10 and maximum number of function evaluations (FEs) is 3000×10.

**Table 3.**

Expected Solutions of PSO Variants with Three VL Strategies: Fixed VL Strategy, Iteration-Dependent Adaptive VL Strategy, Proposed State-Based Adaptive VL Strategy.

| Expected Solutions | | $f_2$ | $f_3$ | $f_5$ | $f_6$ |
|---|---|---|---|---|---|
| | Adopted VL strategy | | | | |
| PSO-LDIW | Fixed | 334781 | 9.39 | 1366 | 10.4 |
| | Iteration-dependent | 300127 | 7.46 | 817 | 0.951 |
| | Proposed state-based | **32062** | **4.84** | **649** | **0.772** |
| FDR-PSO | Fixed | 100064 | 12 | 662 | 2.44 |
| | Iteration-dependent | 112.0 | 5.38 | 606 | 1.24 |
| | Proposed state-based | **18.3** | **4.11** | **545** | **1.04** |
| APSO | Fixed | 611 | **0.1** | 19.7 | 2.45 |
| | Iteration-dependent | 163 | 0.298 | 59.2 | 1.74 |
| | Proposed state-based | **88.2** | 0.117 | **13.7** | **1.27** |
| RODD-PSO | Fixed | 403 | 6.03 | 585 | 0.642 |
| | Iteration-dependent | 367 | 5.48 | 637 | 0.410 |
| | Proposed state-based | **335** | **3.07** | **472** | **0.368** |

In terms of the expected optimization solutions as shown in Table 3, the PSO variants with the state-based adaptive VL strategy are generally better than those with the fixed VL or the iteration-dependent adaptive VL strategies. Hence, the proposed state-based adaptive VL strategy has been validated to refine the optimization solutions of PSO variants.

In terms of the expected FEs in obtaining the final solution as listed in Table 4, the variants with the state-based adaptive VL strategy are generally faster than the variants with the fixed VL or the iteration-dependent VL strategies. Thus, the proposed state-based adaptive VL strategy has been validated to accelerate the convergence speed of PSO variants.

**In a word, the major challenge in the existing adaptive VL strategies has been solved with the proposed state-based adaptive VL strategy, which updates VL based on the current searching state of particles.** The better optimization solutions and the faster convergence speed of the PSO variants with the state-based adaptive VL strategy have been experimentally validated.

**Table 4.**

Expected FEs of PSO Variants with Three VL Strategies: Fixed VL Strategy, Iteration-Dependent Adaptive VL Strategy, Proposed State-Based Adaptive VL Strategy.

| Expected FEs | | $f_2$ | $f_3$ | $f_5$ | $f_6$ |
|---|---|---|---|---|---|
| | Adopted VL strategy | | | | |
| PSO-LDIW | Fixed | 17331 | 23183 | 9846 | 22252 |
| | Iteration-dependent | 16887 | 18885 | **6430** | 20792 |
| | Proposed state-based | **15589** | **16811** | 6713 | **18622** |
| FDR-PSO | Fixed | 9405 | 12135 | 3577 | 10503 |
| | Iteration-dependent | 9938 | 11738 | 2688 | 9272 |
| | Proposed state-based | **7135** | **9519** | **1640** | **8428** |
| APSO | Fixed | 8010 | 16393 | 11379 | 10575 |
| | Iteration-dependent | 6272 | 15769 | 14102 | 8524 |
| | Proposed state-based | **4755** | **15275** | **11205** | **6691** |
| RODD-PSO | Fixed | 12351 | 11608 | 20737 | 12340 |
| | Iteration-dependent | 13779 | 11156 | 6629 | 12975 |
| | Proposed state-based | **9102** | **9440** | **6014** | **9777** |

### 4.3 The Limit Handling Strategies

In PSO-SAVL, the limit handling strategies in [27] are modified by integrating with ESE, which adjust the positions and the velocities of particles to be inside their limits whenever the limits are exceeded based on the value of evolutionary factor $f$.

#### 4.3.1 Velocity Limit Handling Strategy

In PSO-SAVL, the velocity limit handling strategy is combined with ESE, as shown in Algorithm 1, where *Rand* is a uniformly distributed random number within [0, 1]. When the velocity $V_i$ of the $i^{th}$ particle exceeds [-$VL$, $VL$], $V_i$ is either clamped to the nearest limit when $f \in [0.5, 1]$ (global searching) to prevent particles from searching outside the solution space, or randomly distributed within [-$VL$, $VL$] when $f \in [0, 0.5)$ (local searching) to avoid premature convergence.

**Algorithm 1.**

**Velocity Limit Handling.**

```
Algorithm: Velocity Limit Handling
Input: D, V_i, VL, f
Output: V_i
1   FOR d = 1 to D:
2       IF   V_i^d > VL^d or V_i^d < -VL^d:
3           IF f >= 0.5:
4               V_i^d = min(VL^d, max(-VL^d, V_i^d))
5           ELSE:
6               V_i^d = Rand • 2VL^d - VL^d
7   RETURN   V_i
```

### 4.3.2 Position Limit Handling Strategy

**Algorithm 2.**

**Position Limit Handling.**

```
Algorithm: Position Limit Handling
Input: D, X_i, X_min, X_max
Output: X_i
1   FOR d = 1 to D:
2       IF   X_i^d > X_max^d or X_i^d < X_min^d:
3           X_i^d = Rand • (X_max^d - X_min^d) + X_min^d
4   RETURN   X_i
```

As shown in Algorithm 2, with the position limit handling strategy, when the position $X_i$ of the $i^{th}$ particle exceeds [$X_{min}$, $X_{max}$], instead of being clamped to the nearest limits, $X_i$ is randomly distributed within [$X_{min}$, $X_{max}$]. Since the position of particles is randomly distributed when the position limit is exceeded, PSO-SAVL is encouraged to explore the entire solution space, and thus the capability to avoid local optima is improved.

With the limit handling strategies shown in Algorithm 1 and Algorithm 2, the proposed PSO-SAVL can improve the capability of avoiding local optima.

### 4.4 Complexity of the Proposed PSO-SAVL

Based on the flowchart of the proposed PSO-SAVL in Fig. 3, the total time complexity of PSO-SAVL

is $O(D \cdot N^2 \cdot max\_iters)$, where $D$ is the dimension of the problem space, $N$ is the number of particles and $max\_iters$ is the number of iterations. The major parts of the time complexity are attributable to the following steps: the computation of evolutionary factor $f$, the velocity and position update, and the velocity limit and position limit handling algorithms.

The complexity of the computation of evolutionary factor $f$ is $O(D \cdot N^2 \cdot max\_iters)$. As shown in equation (3), the computation of the mean distance from the $i^{th}$ particle to others $d_i$ includes the sum over all dimensions $D$ and the average over all other particles $N$-1, so it requires $O(D \cdot N)$. And for all $N$ particles, $d_i$ should be computed, so the complexity is $O(D \cdot N^2)$. Computation of $d_i$ of all $N$ particles are required for all $max\_iters$ number of iterations, so its total complexity is $O(D \cdot N^2 \cdot max\_iters)$.

The velocity and position update are demanded for all $D$ dimensions of all $N$ particles in all $max\_iters$ iterations, so its time complexity is $O(D \cdot N \cdot max\_iters)$.

The complexity of velocity limit and position limit handling algorithms is $O(D \cdot N \cdot max\_iters)$ in the worst cases. The worst cases occur when the velocity and position of all dimensions of all particles exceed the corresponding limits, so in the worst cases it takes $O(D \cdot N \cdot max\_iters)$ to implement.

In a word, the total time complexity of the proposed PSO-SAVL is $O(D \cdot N^2 \cdot max\_iters)$, which is dominated by the step of calculating the evolutionary factor.

## 5. Experimental Results

In this section, to validate the good performance of the proposed PSO-SAVL, other popular PSO variants are compared with it on the benchmark functions in Table 2 with 50 dimensions. Moreover, the satisfactory scalability of PSO-SAVL is verified on high-dimension and large-scale problems.

### 5.1 Experimental Results on Benchmark Functions with 50 Dimensions
#### 5.1.1 Experimental Configuration

In the experiments, the configuration of the implemented PSO variants is listed in Table 5. The configuration of the proposed PSO-SAVL is as follows. The inertia weight $\omega$ is initialized to 0.9, and is linearly decreased to 0.4, the acceleration coefficients $c_1$ and $c_2$ are 2.05. $\mu_{min}$ and $\mu_{max}$ are selected as 0.4 and 0.7, respectively, the selection of which is discussed in Section 6.2.

For a fair comparison among the PSO variants, the population size $N$ is 20, the value of which is commonly utilized in PSO variants [20]. The dimension $D$ of all the benchmark functions is 50. $max\_iters$ is 10000, so the maximum number of FEs is 20×10000. All the experiments are carried out on the same machine with an Intel i5-8250U CPU and Windows 10 operating system. To reduce the statistical errors, each function independently runs 30 times. The expected solutions, success ratio and expected computation speed of each PSO variant on every benchmark function are correspondingly summarized in Table 6, Table 7 and Table 8.

**Table 5.**

Configuration of Other Implemented PSO Variants.

| PSO variants | Year | Settings | Reference |
|---|---|---|---|
| PSO-LDIW | 1999 | $\omega$: 0.9-0.4; $c_1 = c_2 = 2.05$ | [44] |
| UPSO | 2004 | $\chi = 0.72984$; $c_1 = c_2 = 2.05$; $u = 0.2$; $m = 1$ | [47] |
| HPSO-TVAC | 2004 | $\omega$: 0.9-0.4; $c_1$: 0.75-2.5; $c_2$: 2.5-0.75 | [14] |
| FDR-PSO | 2003 | $c_1 = c_2 = 1.025$; $c_{fdr} = 2.05$ | [46] |
| QPSO | 2004 | g = 0.96 | [48] |
| APSO | 2009 | $\sigma_{max} = 1.0$; $\sigma_{min} = 0.1$ | [15] |
| CLPSO | 2006 | $\omega$: 0.9-0.4; $c = 1.49445$; $m = 7$ | [19] |
| PSO-CL-pbest | 2011 | $\omega$: 0.9-0.4; $c = 1.49445$; $m = 7$ | [49] |
| RODD-PSO | 2019 | $\omega$: 0.9-0.4; $c_1 = c_3$: 0.75-2.5; $c_2 = c_4$: 2.5-0.75; $N_{delay} = 100$ | [20] |
| PPPSO | 2020 | $\omega$: 0.9-0.4; $c_1=2.05$; $c_2=2.05$ | [22] |
| AGLD-PSO | 2020 | $c_1=1.0$; $c_2=0.1$; $M \in [10, \sqrt{N}]$ | [16] |

### 5.1.2 Comparisons on the Expected Solutions

The results of all the PSO variants in terms of the *mean* (expected solutions) and the *std* (standard deviations) are shown in Table 6. The last row of Table 6 shows the rankings of the proposed PSO-SAVL among all the PSO variants by the performance of expected solutions. Fig. 6 graphically presents the convergence characteristics of all the PSO variants on all the benchmark functions.

**Table 6.**

Optimization Solutions of the PSO Variants on 50-Dimension Benchmark Functions.

| Solutions | | $f_1$ | $f_2$ | $f_3$ | $f_4$ | $f_5$ | $f_6$ | $f_7$ |
|---|---|---|---|---|---|---|---|---|
| PSO-LDIW | *mean* | 1.470 | 2.69E+05 | 258.3 | 66.3 | 8502.7 | 55.6 | 174.6 |
| | *std* | 2.484 | 4.48E+05 | 34.4 | 81.9 | 1087.5 | 19.8 | 46.9 |
| UPSO | *mean* | 4.97E-46 | 1082.3 | 117.9 | 0.3523 | 9015.7 | 0.9950 | 45.6 |
| | *std* | 2.72E-45 | 2564.6 | 22.2 | 0.6610 | 1163.0 | 0.4366 | 19.0 |
| HPSO-TVAC | *mean* | 1.52E-13 | 1723.4 | 112.0 | 0.0256 | 4641.8 | 2.13 | 111.8 |
| | *std* | 3.05E-13 | 3551.7 | 35.4 | 0.0311 | 880.4 | 0.9889 | 23.9 |
| FDR-PSO | mean | 1.19E-08 | 874.6 | 192.6 | 21.0 | 4926.9 | 21.1 | 130.8 |
| | std | 3.33E-08 | 1828.5 | 44.5 | 38.7 | 886.2 | 17.1 | 25.9 |
| QPSO | mean | 1.74E-53 | 147.2 | 102.6 | 0.0313 | 4416.8 | 3.19 | 83.2 |
| | std | 6.43E-53 | 174.4 | 33.0 | 0.0383 | 710.6 | 4.54 | 19.7 |
| APSO | mean | 0.285 | 829.7 | 1.6 | 0.2000 | 245.9 | 1.49 | 80.0 |
| | std | 0.766 | 1885.9 | 1.2 | 0.2400 | 164.8 | 0.7811 | 24.2 |
| CLPSO | mean | 4.10E-32 | 97.5 | 139.7 | 1.57E-09 | 6257.3 | 2.53 | 126.9 |
| | std | 4.87E-32 | 55.8 | 15.6 | 6.81E-09 | 436.4 | 0.765 | 16.9 |
| PSO-CL-pbest | mean | 3.77E-39 | 91.8 | 75.4 | 1.15E-05 | 6827.0 | 1.31 | 100.2 |
| | std | 8.39E-39 | 42.4 | 11.1 | 6.31E-05 | 465.9 | 0.204 | 16.4 |
| RODD-PSO | mean | 2.85E-43 | 155.0 | 139.2 | 0.0316 | 4302.9 | 1.87 | 88.3 |
| | std | 1.56E-42 | 181.2 | 47.9 | 0.0353 | 862.5 | 0.6499 | 20.2 |
| PPPSO | mean | 6.65E-34 | 246.1 | 82.1 | 0.0163 | 6246.3 | 1.86 | 45.0 |
| | std | 1.47E-33 | 115.4 | 50.2 | 0.0101 | 1126.3 | 0.4918 | 23.6 |
| AGLD-PSO | mean | 1.91E-38 | 128.9 | 99.6 | 2.73E-03 | 4355.2 | 1.47 | 58.9 |
| | std | 7.12E-38 | 102.2 | 35.7 | 5.64E-03 | 1099.4 | 0.6072 | 23.2 |
| PSO-SAVL | mean | 1.08E-39 | 76.6 | 55.8 | 0.0107 | 4172.7 | 0.9902 | 30.8 |
| | std | 1.54E-39 | 39.3 | 13.5 | 0.0185 | 901.1 | 0.0970 | 8.2 |
| **Rankings of PSO-SAVL** | | 4 | 1 | 2 | 4 | 2 | 1 | 1 |

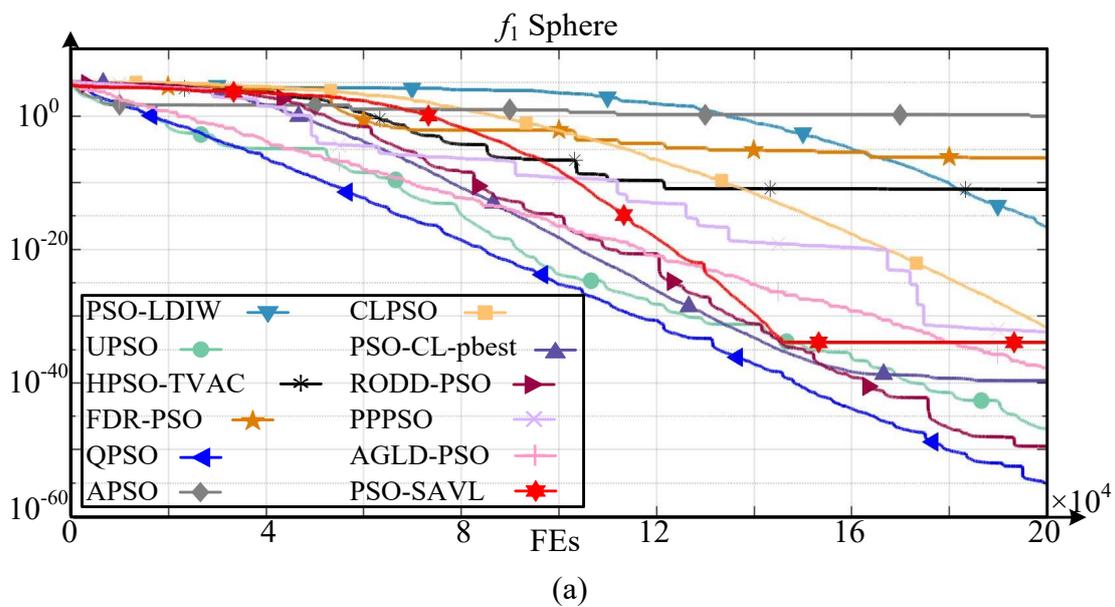

(a)

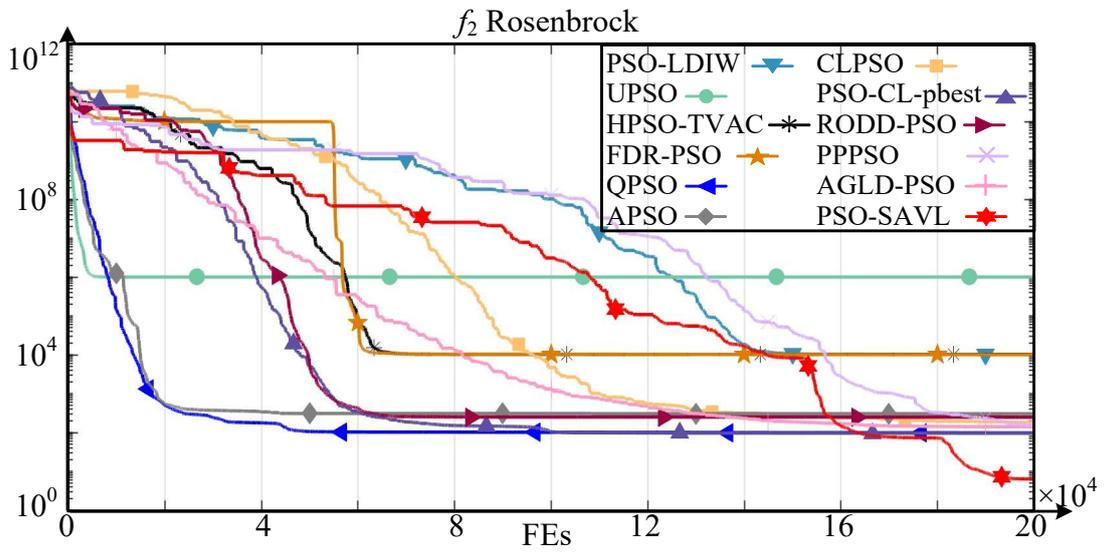

(b)

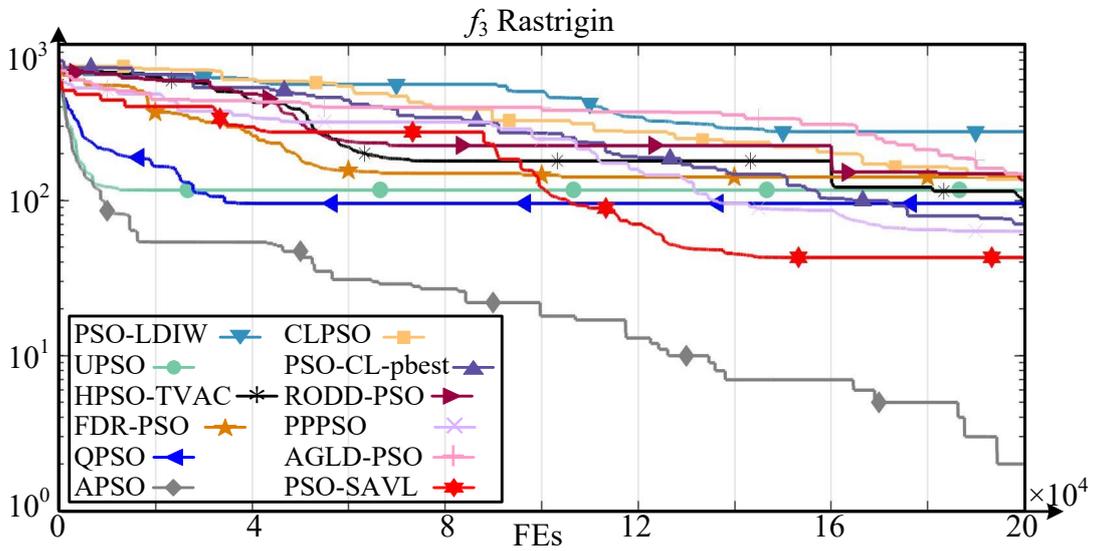

(c)

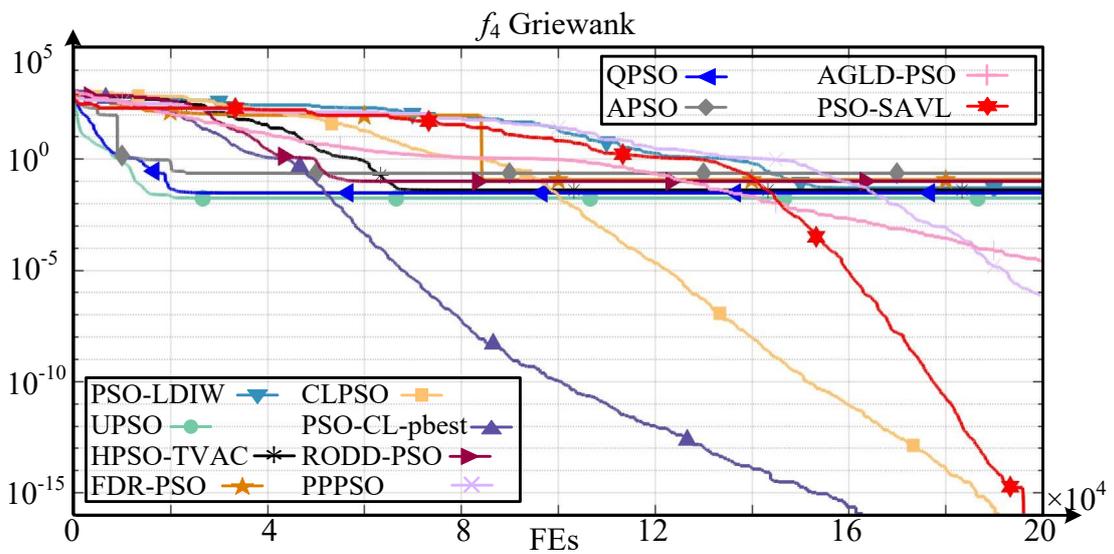

(d)

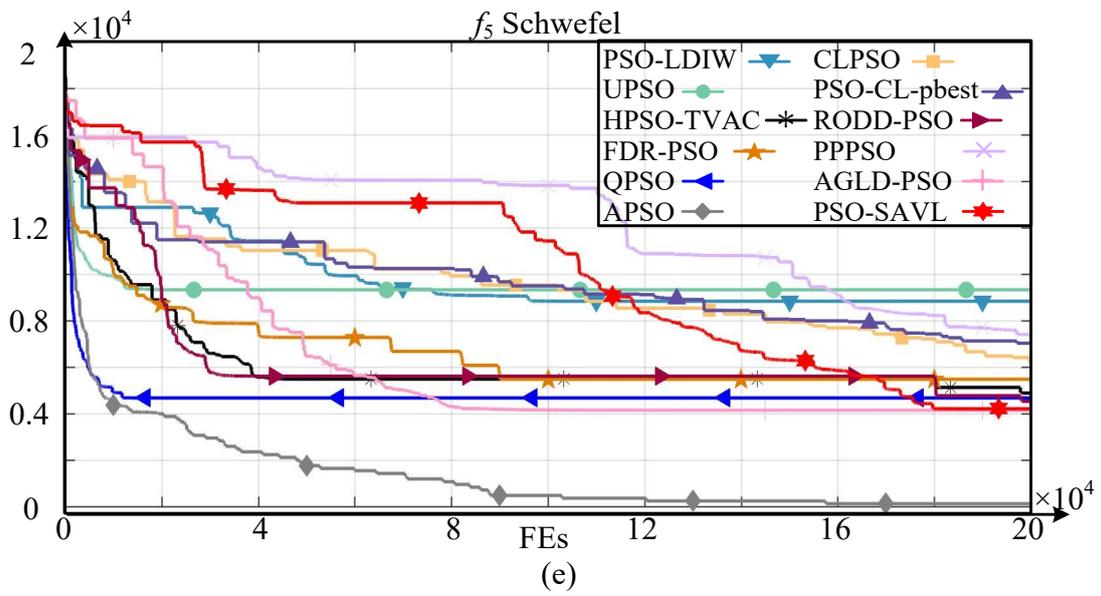

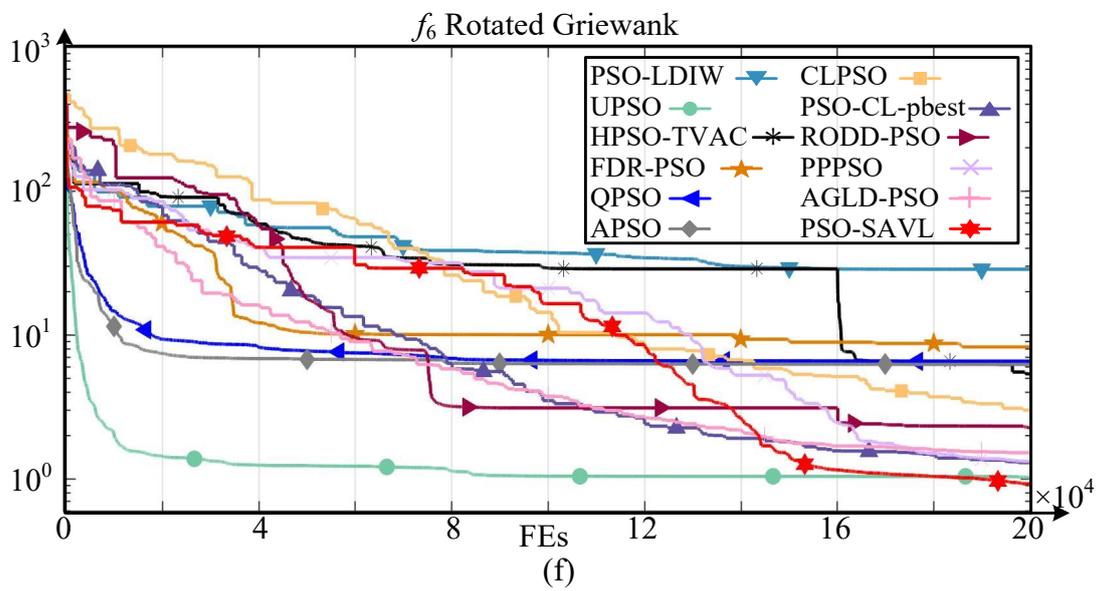

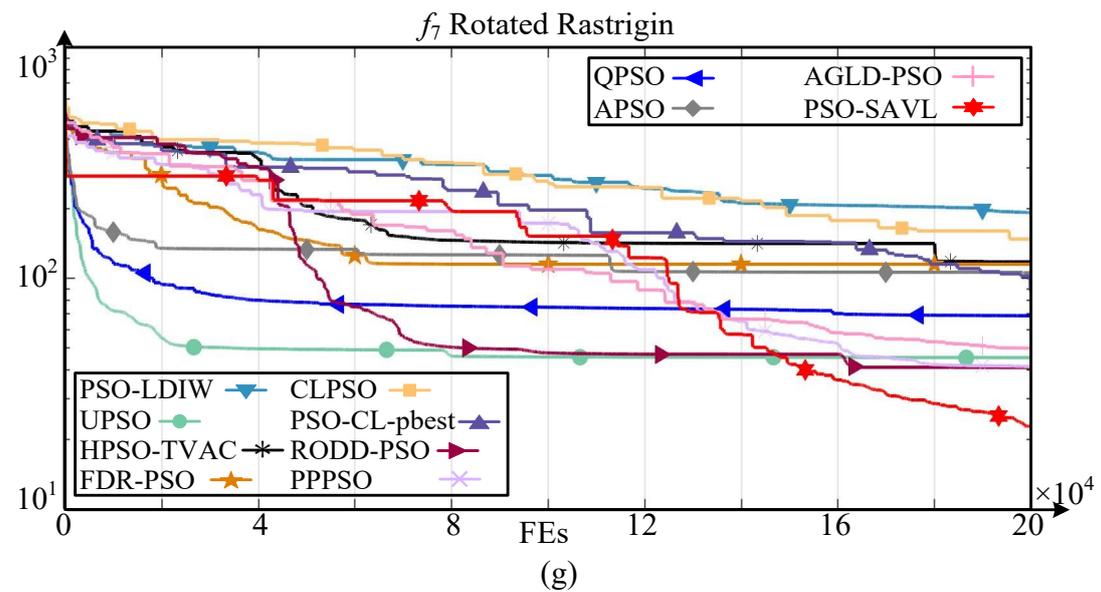

**Fig. 6.** Convergence performance of all the PSO variants on each of the benchmark functions: (a) $f_1$; (b) $f_2$; (c) $f_3$; (d) $f_4$; (e) $f_5$; (f) $f_6$; (g) $f_7$.

Generally, the results of the expected solutions in Table 6 show that the proposed PSO-SAVL achieves much better expected solutions on most of the test functions than PSO-LDIW, UPSO, HPSO-TVAC, FDR-PSO and QPSO. And it reaches satisfactory expected solutions when compared with APSO, CLPSO and PSO-CL-pbest, all of which are famous for their global searching capability. When compared with the state-of-the-art PSO variants including RODD-PSO, PPPSO and AGLD-PSO, the proposed PSO-SAVL also manifests better or comparable expected solutions.

To be more specific, in terms of the unimodal function $f_1$, PSO-SAVL ranks $4^{th}$ among all 12 PSO variants, and is worse than UPSO, QPSO and RODD-PSO. As for the unrotated multimodal function $f_4$, PSO-SAVL ranks $4^{th}$, and is only worse than CLPSO, PSO-CL-pbest and AGLD-PSO. In terms of the unrotated multimodal functions $f_3$ and $f_5$, PSO-SAVL ranks $2^{nd}$, and is better than all the compared PSO variants except for APSO. When solving the complicated rotated multimodal functions $f_6$ and $f_7$ and the unrotated multimodal function $f_2$, PSO-SAVL ranks $1^{st}$ among all the PSO variants, even better than APSO, CLPSO and PSO-CL-pbest, which are famous for their global searching capability, and also better than the cutting-edge RODD-PSO, PPPSO and AGLD-PSO.

In a word, in terms of the expected solutions, the proposed PSO-SAVL shows the best performance on complex rotated multimodal functions. And as for the unrotated multimodal and unimodal functions, it also reaches satisfactory expected optimization solutions. Consequently, the capability of avoiding local optima of the proposed PSO-SAVL has been validated.

### 5.1.3 Comparisons on the Success Ratio

Table 7.

Success Ratio of the PSO Variants on 50-Dimension Benchmark Functions.

| Success Ratio | PSO-LDIW | UPSO | HPSO-TVAC | FDR-PSO | QPSO | APSO |
|---|---|---|---|---|---|---|
| $f_1$ | 56.7% | 100.0% | 100.0% | 100.0% | 100.00% | 66.7% |
| $f_2$ | 36.7% | 86.7% | 76.7% | 70.0% | 93.33% | 70.0% |
| $f_3$ | 0.0% | 0.0% | 0.0% | 0.0% | 3.33% | 100.0% |
| $f_4$ | 53.3% | 86.7% | 100.0% | 76.7% | 100.00% | 90.0% |
| $f_5$ | 13.3% | 3.3% | 100.0% | 100.0% | 100.00% | 100.0% |
| $f_6$ | 0.0% | 100.0% | 100.0% | 10.0% | 83.33% | 100.0% |
| $f_7$ | 26.7% | 100.0% | 93.3% | 80.0% | 100.00% | 100.0% |
| **Mean Success Ratio** | 26.7% | 68.1% | 81.4% | 62.4% | 82.86% | 89.5% |
| **Success Ratio** | CLPSO | PSO-CL-pbest | RODD-PSO | PPPSO | AGLD-PSO | PSO-SAVL |
| $f_1$ | 100.0% | 100.0% | 100.0% | 100.0% | 100.0% | 100.0% |
| $f_2$ | 100.0% | 100.0% | 93.3% | 83.3% | 90.0% | 100.0% |
| $f_3$ | 0.0% | 0.0% | 26.7% | 36.7% | 30.0% | 33.3% |
| $f_4$ | 100.0% | 100.0% | 100.0% | 100.0% | 100.0% | 100.0% |
| $f_5$ | 96.7% | 70.0% | 100.0% | 100.0% | 100.0% | 100.0% |
| $f_6$ | 100.0% | 100.0% | 100.0% | 96.7% | 100.0% | 100.0% |
| $f_7$ | 93.3% | 100.0% | 3.3% | 100.0% | 100.0% | 100.0% |
| **Mean Success Ratio** | 84.3% | 81.4% | 74.8% | 88.1% | 88.6% | **90.5%** |

The results regarding success ratio of 12 PSO variants are listed in Table 7, which is the ratio of the number of solutions that are lower than the "Acceptance" threshold in Table 2 to the number of repetitions which is 30. The proposed PSO-SAVL achieves 100% success ratio on all functions except for the function $f_3$, on which only APSO reaches 100%. In comparison to the latest PSO variants, RODD-PSO, PPPSO and AGLD-PSO, PSO-SAVL also realizes higher mean success ratio. Among all 12 PSO variants, PSO-SAVL offers the highest mean success ratio of 90.48%, 0.96% higher than the 2$^{nd}$ APSO. In summary, from the perspective of success ratio, the proposed PSO-SAVL displays outstanding performance.

### 5.1.4  Comparisons on the Expected Computation Speed

**Table 8.**

Computation Speed of the PSO Variants on 50-Dimension Benchmark Functions.

| **Expected Time** | PSO-LDIW | UPSO | HPSO-TVAC | FDR-PSO | QPSO | APSO |
|---|---|---|---|---|---|---|
| $f_1$ | 3.10 | 4.48 | 3.75 | 10.52 | 4.06 | 5.68 |
| $f_2$ | 3.37 | 4.43 | 4.04 | 10.73 | 4.58 | 6.06 |
| $f_3$ | 3.88 | 4.79 | 4.22 | 10.68 | 4.88 | 6.76 |
| $f_4$ | 4.23 | 4.69 | 3.98 | 10.44 | 4.41 | 6.44 |
| $f_5$ | 4.08 | 4.48 | 3.85 | 10.78 | 4.44 | 6.29 |
| $f_6$ | 6.65 | 6.91 | 6.11 | 16.66 | 6.94 | 10.06 |
| $f_7$ | 6.17 | 6.67 | 5.92 | 16.12 | 6.80 | 9.30 |
| **Mean Expected Time** | 4.50 | 5.21 | 4.55 | 12.28 | 5.16 | 7.23 |
| **Expected Time** | CLPSO | PSO-CL-pbest | RODD-PSO | PPPSO | AGLD-PSO | PSO-SAVL |
| $f_1$ | 4.94 | 4.65 | 23.47 | 1.71 | 6.80 | 5.09 |
| $f_2$ | 4.95 | 4.97 | 23.87 | 2.02 | 7.15 | 5.41 |
| $f_3$ | 5.55 | 5.46 | 25.05 | 2.12 | 7.09 | 5.63 |
| $f_4$ | 5.28 | 5.28 | 25.51 | 2.06 | 6.90 | 5.43 |
| $f_5$ | 5.21 | 5.24 | 25.46 | 2.14 | 6.94 | 5.60 |
| $f_6$ | 8.14 | 8.14 | 36.61 | 2.34 | 7.89 | 5.26 |
| $f_7$ | 7.44 | 7.88 | 37.66 | 2.49 | 8.66 | 5.97 |
| **Mean Expected Time** | 5.93 | 5.94 | 28.23 | 2.13 | 7.35 | 5.48 |

In PSO-SAVL, although the state-based adaptive VL strategy and the limit handling strategies require extra operations, Table 8 shows that the computation speed of PSO-SAVL does not slow down and is still satisfactory. Except the fast PPPSO due to its multi-thread parallel strategy [22], PSO-SAVL shows faster or comparable computation speed than other considered PSO variants. Specifically, PSO-SAVL displays faster computation speed than FDR-PSO, APSO, CLPSO, PSO-CL-pbest, RODD-PSO and AGLD-PSO. When compared with PSO-LDIW, UPSO, HPSO-TVAC and QPSO, the proposed PSO-SAVL shows the comparable expected computation time.

### 5.1.5 Comparisons through t-Tests

For a convincing analysis, the two-tailed t-test [15] is carried out. Table 9 presents the $t$ values and the $P$ values on every test function between the proposed PSO-SAVL and another PSO variants with a significance level of 0.05. Rows "Better" and "Worse" represent the number of functions on which PSO-SAVL shows significantly better, and significantly worse performance than the compared PSO variants, respectively. Row "Same" gives the number of functions on which PSO-SAVL displays almost the same performance as the compared PSO variants. Row "General merit" shows the difference between the number of "Better" and the number of "Worse", indicating the overall comparison between PSO-SAVL and the compared variant. In Table 9, the orange and blue blocks show the benchmark functions on which PSO-SAVL is significantly better and significantly worse, respectively.

**Table 9.**

Comparisons between PSO-SAVL and Others through t-Tests on 50-Dimension Benchmark Functions.

|  |  | PSO-LDIW | UPSO | HPSO-TVAC | FDR-PSO | QPSO | APSO |
|---|---|---|---|---|---|---|---|
| $f_1$ | $t$-value | 3.241 | -3.852 | 2.732 | 1.961 | -3.852 | 2.037 |
|  | $P$-value | 1.98E-03 | 2.95E-04 | 8.33E-02 | 5.46E-02 | 2.95E-04 | 4.62E-02 |
| $f_2$ | $t$-value | 3.290 | 2.184 | 2.539 | 2.390 | 2.163 | 2.187 |
|  | $P$-value | 1.71E-03 | 3.30E-02 | 1.38E-02 | 2.01E-02 | 3.46E-02 | 3.28E-02 |
| $f_3$ | $t$-value | 29.996 | 13.073 | 8.117 | 16.104 | 7.183 | -21.978 |
|  | $P$-value | 5.18E-37 | 6.15E-19 | 3.90E-11 | 4.53E-23 | 1.44E-09 | 8.28E-30 |
| $f_4$ | $t$-value | 4.430 | 2.830 | 2.251 | 2.976 | 2.656 | 4.308 |
|  | $P$-value | 4.23E-05 | 6.39E-03 | 2.82E-02 | 4.25E-03 | 1.02E-02 | 6.45E-05 |
| $f_5$ | $t$-value | 16.793 | 18.031 | 2.040 | 3.269 | 1.165 | -23.480 |
|  | $P$-value | 6.08E-24 | 1.89E-25 | 4.59E-02 | 1.82E-03 | 2.49E-01 | 2.62E-31 |
| $f_6$ | $t$-value | 15.072 | 0.060 | 6.284 | 6.443 | 2.649 | 3.495 |
|  | $P$-value | 1.02E-21 | 0.953 | 4.61E-08 | 2.51E-08 | 0.0104 | 9.14E-04 |
| $f_7$ | $t$-value | 16.538 | 3.936 | 17.549 | 20.129 | 13.470 | 10.569 |
|  | $P$-value | 1.27E-23 | 2.25E-04 | 7.15E-25 | 7.57E-28 | 1.65E-19 | 3.83E-15 |
| Better |  | 7 | 5 | 7 | 6 | 5 | 5 |
| Same |  | 0 | 1 | 0 | 1 | 1 | 0 |
| Worse |  | 0 | 1 | 0 | 0 | 1 | 2 |
| **General merit** |  | **7** | **4** | **7** | **6** | **4** | **3** |

|  |  | CLPSO | PSO-CL-pbest | RODD-PSO | PPPSO | AGLD-PSO |
|---|---|---|---|---|---|---|
| $f_1$ | $t$-value | 4.610 | 1.725 | -3.851 | 2.478 | 1.386 |
|  | $P$-value | 2.26E-05 | 8.99E-02 | 2.96E-04 | 1.62E-02 | 1.71E-01 |
| $f_2$ | $t$-value | 1.674 | 1.438 | 2.315 | 7.614 | 2.616 |
|  | $P$-value | 9.96E-02 | 1.56E-01 | 2.42E-02 | 2.71E-10 | 1.13E-02 |
| $f_3$ | $t$-value | 22.321 | 6.126 | 9.181 | 2.771 | 6.286 |
|  | $P$-value | 3.71E-30 | 8.43E-08 | 6.66E-13 | 7.50E-03 | 4.59E-08 |
| $f_4$ | $t$-value | -3.176 | -3.172 | 2.863 | 1.455 | -2.257 |
|  | $P$-value | 2.39E-03 | 2.42E-03 | 5.83E-03 | 1.51E-01 | 2.78E-02 |
| $f_5$ | $t$-value | 11.405 | 14.332 | 0.572 | 7.874 | 0.703 |
|  | $P$-value | 1.90E-16 | 1.03E-20 | 5.70E-01 | 9.96E-11 | 4.85E-01 |
| $f_6$ | $t$-value | 10.911 | 7.665 | 7.312 | 9.504 | 4.274 |
|  | $P$-value | 1.11E-15 | 2.23E-10 | 8.75E-10 | 1.98E-13 | 7.24E-05 |
| $f_7$ | $t$-value | 28.076 | 20.689 | 14.469 | 3.113 | 6.255 |
|  | $P$-value | 1.88E-35 | 1.87E-28 | 6.66E-21 | 2.87E-03 | 5.16E-08 |
| Better |  | 5 | 4 | 5 | 6 | 4 |
| Same |  | 1 | 2 | 1 | 1 | 2 |
| Worse |  | 1 | 1 | 1 | 0 | 1 |
| **General merit** |  | **4** | **3** | **4** | **6** | **3** |

\*: The statistical t-Test determines whether there is a significant difference between the solutions of the proposed PSO-SAVL and the compared PSO variant [15], and the degree of freedom and the significance level of the t-Tests are 29 and 0.05, respectively.

☐ : The benchmark functions on which the proposed PSO-SAVL is significantly better than the compared PSO variant.

☐ : The benchmark functions on which the proposed PSO-SAVL is significantly worse than the compared PSO variant.

The proposed PSO-SAVL significantly outperforms PSO-LDIW, HPSO-TVAC and FDR-PSO on most of the benchmark functions, because the "General merit" of the comparison between PSO-SAVL and each of the three variants is larger than or equal to 6. Moreover, PSO-SAVL is obviously better than UPSO, QPSO and CLPSO, since the "General merit" of the comparison between PSO-SAVL and each of the five variants is 4. When compared with APSO and PSO-CL-pbest, PSO-SAVL achieves "General merit" of 3, representing the better optimization results of the proposed PSO-SAVL.

Compared with the latest cutting-edge RODD-PSO, PPPSO and AGLD-PSO, the proposed PSO-SAVL is also better. As shown in Table 9, the "General Merit" of PSO-SAVL compared with PPPSO is 6, indicating the significant better optimization performance of PSO-SAVL. Moreover, the "General Merit" of the comparison between PSO-SAVL and RODD-PSO and AGLD-PSO is 3 and 4, respectively, indicating better results of PSO-SAVL.

To conclude the overall performance of PSO-SAVL on benchmark functions with 50 dimensions, in terms of the expected solutions, the proposed PSO-SAVL realizes the best expected solutions on rotated multimodal functions, and it reaches satisfactory expected solutions on unimodal and unrotated multimodal functions. From the perspective of computation speed and success ratio, PSO-SAVL realizes the highest success ratio in average, and it keeps the fast computation speed among all the variants considered. The statistical t-test further provides a convincing analysis about the performance of PSO-SAVL when compared with other PSO variants.

## 5.2 Scalability of PSO-SAVL on High-Dimension and Large-Scale Problems

To study the scalability of the proposed PSO-SAVL on high-dimension and large-scale problems, the considered PSO variants in Table 5 are compared with PSO-SAVL on the multimodal function $f_2$ and the rotated multimodal functions $f_6$ and $f_7$. On the high-dimension and large-scale problems, the number of particles $N$ is $0.5D$, the maximum number of iterations $max\_iters$ is 10000, so the maximum number of function evaluations is $5000D$. The settings of PSO variants are the same as those in Table 5.

The expected solutions of 30 independent repetitions of all PSO variants on $f_2$, $f_6$ and $f_7$ are shown in Fig. 7 (a), Fig. 7 (b) and Fig. 7 (c), respectively. Based on Fig. 7, as the dimension $D$ increases to 200 and 350 (high-dimension-scale), the performance of many PSO variants deteriorates, such as PSO-LDIW, UPSO, QPSO and PSO-CL-pbest. Despite that, the performance of PSO-SAVL in high-dimension problems is satisfactory: PSO-SAVL ranks 3$^{rd}$, 1$^{st}$, 1$^{st}$ on $f_2$, $f_6$ and $f_7$ with 200 dimensions, and it ranks 2$^{nd}$, 1$^{st}$, 2$^{nd}$ on $f_2$, $f_6$ and $f_7$ with 350 dimensions.

From the perspective of large-scale problems (500 dimensions), PSO-SAVL reaches the 3$^{rd}$ and is only worse than CLPSO and PPPSO on $f_2$. It ranks 2$^{nd}$ and is slightly worse than RODD-PSO on $f_6$. It also ranks 2$^{nd}$ and is only worse than UPSO on $f_7$. Apart from that, in comparison to the state-of-the-art AGLD-PSO on the benchmark functions of 500 dimensions as shown in Table 10, PSO-SAVL is slightly better on $f_2$, $f_3$, $f_6$ and $f_7$, and shows comparable optimization results on $f_1$, $f_4$ and $f_5$.

To conclude, in high-dimension and large-scale problems, the proposed PSO-SAVL can still achieve good optimization results, and its scalability has been empirically validated.

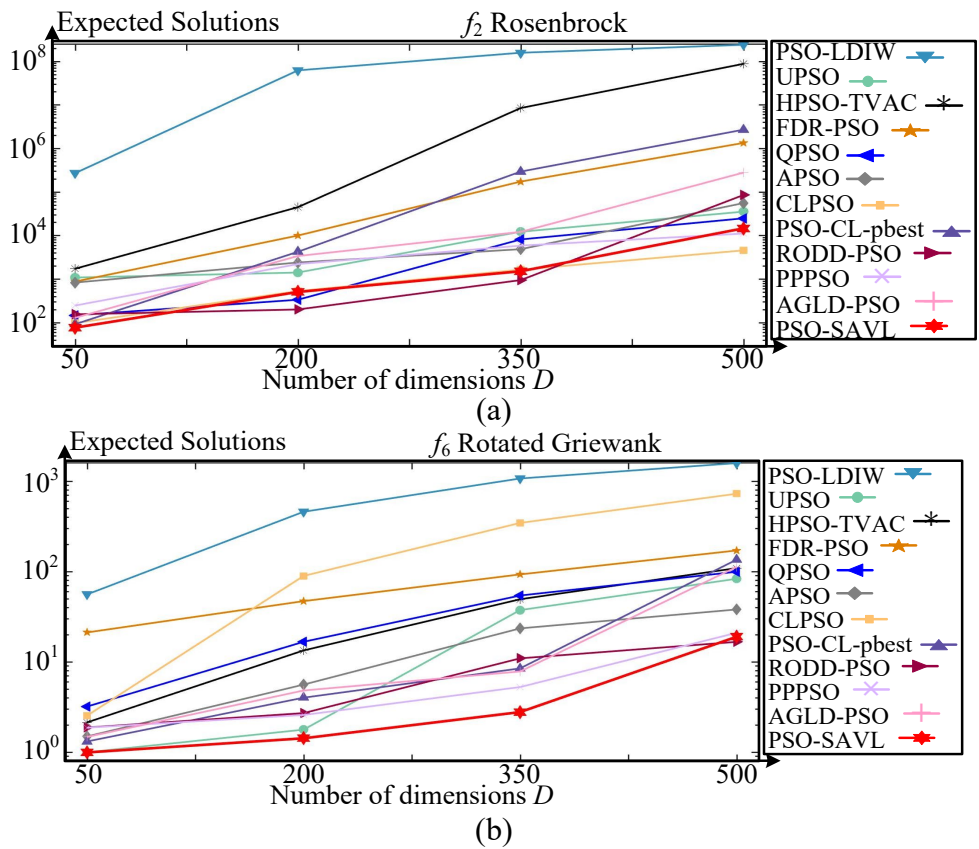

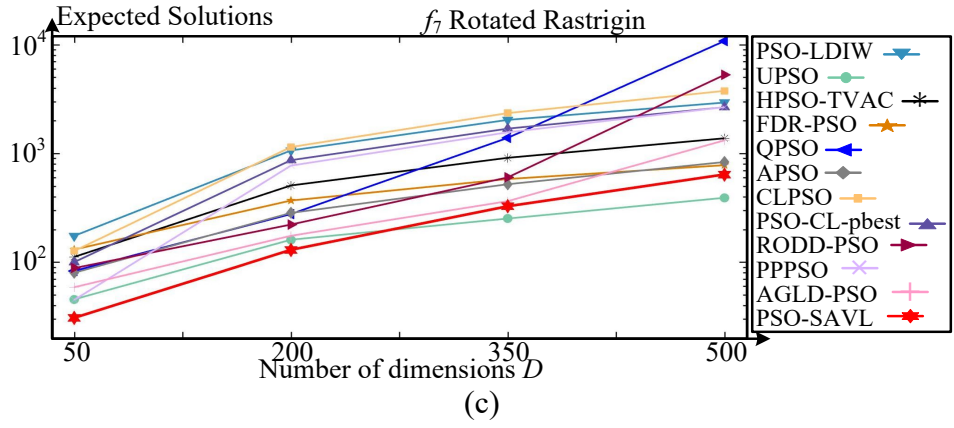

**Fig. 7.** Expected solutions of all PSO variants in high-dimension and large-scale problems: (a) $f_2$; (b) $f_6$; (c) $f_7$.

**Table 10.**

Expected Solutions of AGLD-PSO and PSO-SAVL on 500-D Large-Scale Problems.

| **Expected Solutions** | $f_1$ | $f_2$ | $f_3$ | $f_4$ |
|---|---|---|---|---|
| AGLD-PSO | **3.68E-05** | 2.78E+05 | 6.20E+03 | **3.33E+03** |
| PSO-SAVL | 1.96E-04 | **1.46E+04** | **4.04E+03** | 1.03E+04 |
| **Expected Solutions** | $f_5$ | | $f_6$ | $f_7$ |
| AGLD-PSO | **1.90E+05** | | 1.12E+02 | 1.32E+03 |
| PSO-SAVL | 3.08E+05 | | **1.89E+01** | **6.41E+02** |

## 6. Merits of the State-Based Adaptive VL Strategy and the Limit Handling Strategies in PSO-SAVL and Sensitivity Analysis of Hyper-Parameters

In Section 6.1, the merits of the state-based adaptive VL strategy and the limit handling strategies in the proposed PSO-SAVL are analyzed. In Section 6.2, sensitivity analysis of the newly-introduced hyper-parameters $\mu_{min}$ and $\mu_{max}$ in PSO-SAVL is studied, and how to select $\mu_{min}$ and $\mu_{max}$ is thrown light on.

### 6.1 Merits of the State-Based Adaptive VL Strategy and the Limit Handling Strategies in PSO-SAVL

To elaborate the merits of the strategies in PSO-SAVL, the PSO-SAVL with or without the state-based adaptive VL strategy or the limit handling strategies is tested with the same experimental configuration as in Section 5, and the results in terms of the expected solutions are summarized in Table 11.

In the comparison between the PSO-SAVL without both strategies and the PSO-SAVL with only the state-based adaptive VL strategy, the one with only the state-based adaptive VL strategy achieves better expected solutions on all the benchmark functions. However, its expected solutions are still not satisfying when compared with those of the popular PSO variants APSO, CLPSO, RODD-PSO, PPPSO and AGLD-PSO whose expected solutions are shown in Table 6.

**Table 11.**

Merits of the State-Based Adaptive VL Strategy and the Limit Handling Strategies in the Proposed PSO-SAVL.

| Expected Solutions | Without Both | With only State-Based Adaptive VL | With only Limit Handling Strategies | With Both |
|---|---|---|---|---|
| $f_1$ | 1.573 | 7.37E-36 | 1.06E-34 | **1.08E-39** |
| $f_2$ | 2.69E+05 | 1.56E+05 | 95.6 | **88.6** |
| $f_3$ | 258.3 | 157.1 | 63.9 | **55.8** |
| $f_4$ | 66.3 | 18.1 | **0.00851** | 0.0129 |
| $f_5$ | 8502.7 | 6686.3 | 4812.8 | **4172.7** |
| $f_6$ | 55.6 | 17.6 | 1.073 | **0.9902** |
| $f_7$ | 174.6 | 117.4 | 35.9 | **30.8** |

When compared with the PSO-SAVL with only the state-based adaptive VL strategy, the PSO-SAVL with only the limit handling strategies realizes much better expected solutions, and it shows comparable expected solutions in comparison with the popular PSO variants.

However, the PSO-SAVL with both of the state-based adaptive VL strategy and the limit handling strategies generally outperforms the PSO-SAVL with only the limit handling strategies, because it can adequately update VL to match the current searching state of particles to refine the expected solutions. Since the PSO-SAVL with both strategies achieves the best performance on almost all the functions among the four variants of PSO-SAVL in Table 11, it is reasonable to conclude that both of the state-based adaptive VL strategy and the limit handling strategies are significant in ensuring the best performance of PSO-SAVL.

## 6.2 Sensitivity Analysis of Hyper-Parameters in PSO-SAVL

**Table 12.**

Sensitivity of $\mu_{max}$: Expected Solutions of PSO-SAVL with Different Values of $\mu_{max}$ ($\mu_{min}=0.4$).

| Expected Solutions | $f_1$ | $f_2$ | $f_3$ | $f_4$ | $f_5$ | $f_6$ | $f_7$ |
|---|---|---|---|---|---|---|---|
| $\mu_{max}=1.0$ | 2.60E-39 | 112.67 | 62.09 | 1.22E-02 | 4297.29 | 1.11 | 33.22 |
| $\mu_{max}=0.9$ | 2.72E-39 | 96.48 | 60.84 | 1.38E-02 | **3994.11** | 1.10 | 33.61 |
| $\mu_{max}=0.8$ | 2.14E-39 | 96.05 | 59.61 | **7.02E-03** | 4336.93 | 1.04 | 31.44 |
| $\mu_{max}=0.7$ | **1.08E-39** | **76.6** | **55.8** | 1.07E-02 | 4172.7 | **0.9902** | 30.8 |
| $\mu_{max}=0.6$ | 1.22E-39 | 95.36 | 59.15 | 1.30E-02 | 4373.16 | 1.07 | **30.54** |
| $\mu_{max}=0.5$ | 1.18E-39 | 113.40 | 55.79 | 1.44E-02 | 4298.14 | 1.02 | 31.66 |
| $\mu_{max}=0.4$ | 1.37E-39 | 112.97 | 57.93 | 1.56E-02 | 4396.85 | 1.08 | 32.61 |

**Table 13.**

Sensitivity of $\mu_{min}$: Expected Solutions of PSO-SAVL with Different Values of $\mu_{min}$ ($\mu_{max} = 0.7$).

| Expected Solutions | $f_1$ | $f_2$ | $f_3$ | $f_4$ | $f_5$ | $f_6$ | $f_7$ |
|---|---|---|---|---|---|---|---|
| $\mu_{min} = 0.1$ | 8.85E-18 | 1.20E+09 | 140.4 | 3.60E+01 | 9294.35 | 22.30 | 150.04 |
| $\mu_{min} = 0.2$ | 7.63E-36 | 736.07 | 57.7 | 2.80E-02 | 4545.32 | 4.78 | 31.07 |
| $\mu_{min} = 0.3$ | **5.20E-40** | 119.53 | **54.6** | 1.08E-03 | **3960.65** | 1.10 | **26.31** |
| $\mu_{min} = 0.4$ | 1.08E-39 | **76.7** | 55.8 | **1.07E-02** | 4172.7 | **0.9902** | 30.8 |
| $\mu_{min} = 0.5$ | 1.25E-39 | 112.23 | 61.7 | 1.20E-02 | 4359.30 | 1.09 | 30.83 |
| $\mu_{min} = 0.6$ | 9.26E-40 | 94.14 | 61.4 | 1.72E-02 | 4242.83 | 1.03 | 31.06 |
| $\mu_{min} = 0.7$ | 1.78E-39 | 104.37 | 66.0 | 1.60E-02 | 4307.25 | 1.09 | 30.0 |

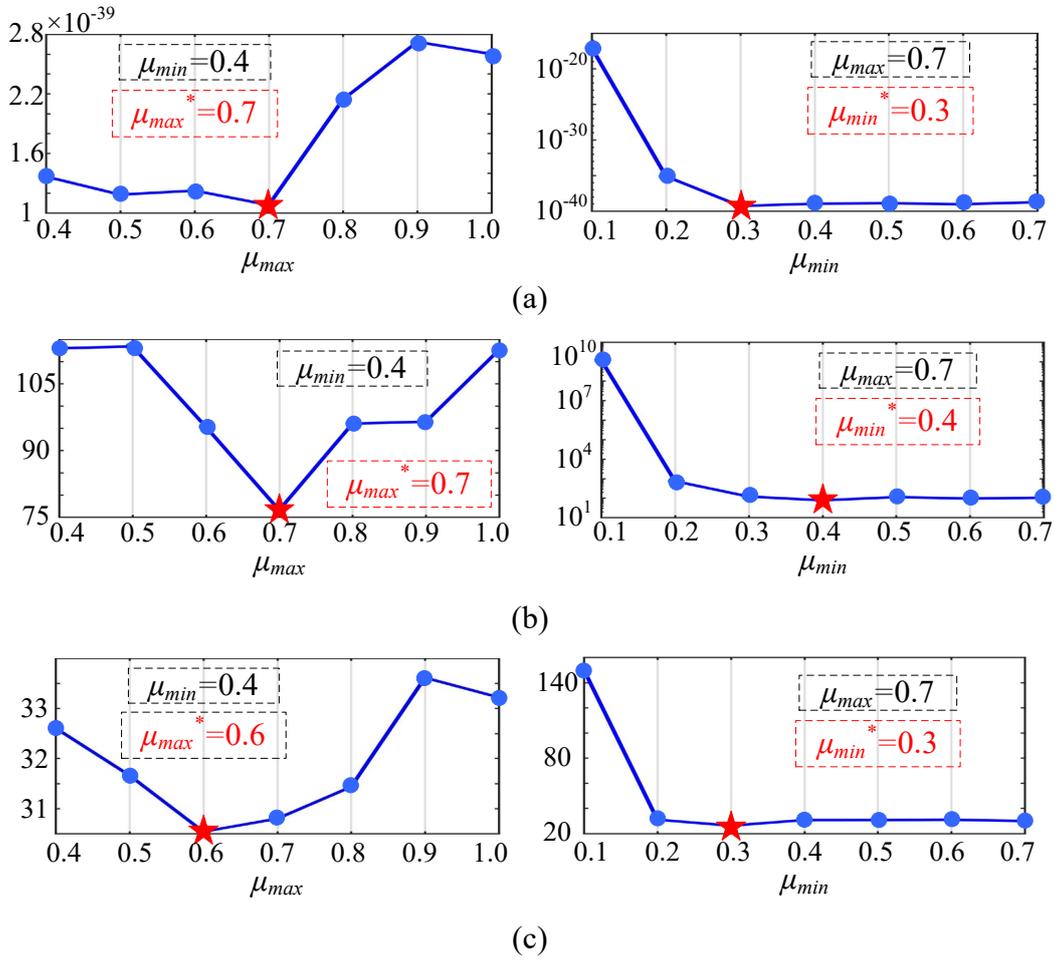

**Fig. 8.** Expected optimization solutions of PSO-SAVL with different values of $\mu_{min}$ and $\mu_{max}$ on functions: (a) Sphere $f_1$; (b) Rosenbrock $f_2$; (c) Rotated-Rastrigin $f_7$.

In the proposed PSO-SAVL, $\mu_{min}$ and $\mu_{max}$ (the minimum and maximum proportions of $VL$ to $X_{max}$) are newly introduced, the sensitivity of which is detailed analyzed here. To elaborate the sensitivity of $\mu_{max}$, at a fixed $\mu_{min}$ at 0.4, the PSO-SAVL with different $\mu_{max}$ within [0.4, 1.0] is tested on the benchmark functions in Table 2 of 50 dimensions, and the results are presented in Table 12 and Fig. 8. To analyze the sensitivity of $\mu_{min}$, at a fixed $\mu_{max}$ at 0.7, the PSO-SAVL with different $\mu_{min}$ within [0.1, 0.7] is tested, and the results are presented in Table 13 and Fig. 8. The results in Table 12, Table 13 and Fig. 8 are the expected solutions found in 30 independent trials.

As can be seen in Table 12, PSO-SAVL is not very sensitive to $\mu_{max}$, and the values of $\mu_{max}$ from 0.4 to 1.0 all offer satisfactory optimization results. Thus, it is reasonable to conclude that the performance of PSO-SAVL is robust against the changing value of $\mu_{max}$. According to the best results in bold font in Table 12 and the left-hand side of Fig. 8, it is recommended that $\mu_{max}$ to be selected within [0.6, 0.8] for the best optimization results. Hence, it is always good to start with 0.7, explaining the $\mu_{max}$ of 0.7 in the proposed PSO-SAVL in Fig. 3.

From the perspective of $\mu_{min}$, the optimization results of PSO-SAVL deteriorate when the value of $\mu_{min}$ is too small. As a result, based on Table 13 and the right-hand side of Fig. 8, small values of $\mu_{min}$, such as 0.1 and 0.2, should be avoided. And values around 0.3 and 0.4 are suitable for the selection of $\mu_{min}$, since these values can result in the best optimization results, which also explains why $\mu_{min}$ is selected as 0.4 in PSO-SAVL.

In a word, PSO-SAVL is robust against changing values of $\mu_{max}$, and too small value of $\mu_{min}$ should be avoided. And it is suggested that $\mu_{max}$ is chosen within [0.6, 0.8] and $\mu_{min}$ is selected within [0.3, 0.4].

## 7. Conclusions

In this paper, to mitigate the incompatibility between the existing adaptive VL strategies and the current state of particles, a state-based adaptive VL strategy is proposed. The proposed state-based adaptive VL strategy tunes VL based on the value of evolutionary factor to match the current searching state of particles, and has been validated to refine the optimization solutions and accelerate the convergence speed of PSO variants.

With the state-based adaptive VL strategy, a novel PSO variant, PSO-SAVL, is proposed, in which limit handling strategies are adopted. The proposed PSO-SAVL is compared with 12 existing PSO variants on a wide range of benchmark functions, including unimodal, unrotated multimodal and rotated multimodal functions. In the experiments of benchmark functions of 50 dimensions, in comparison to the PSO variants tested, the proposed PSO-SAVL achieves better expected solutions on unimodal and unrotated multimodal functions and obtains the best expected solutions on complex rotated multimodal functions. Apart from that, PSO-SAVL ranks 1st in terms of the expected success ratio and has fast computation speed among the PSO variants tested. Furthermore, when PSO-SAVL is applied in high-dimension and large-scale problems, its satisfactory optimization performance has been empirically verified.

Finally, the merits of the strategies in PSO-SAVL have been experimentally validated, proving that both of the state-based adaptive VL strategy and the limit handling strategies are necessary and important in ensuring the best performance of PSO-SAVL. Sensitivity of the newly-introduced hyper-parameters in PSO-SAVL is explicitly studied, and insights in how to select these hyper-parameters is also thrown light on.

To extend the work of PSO with state-based adaptive velocity limit strategy in this paper, the following aspects can be considered. Firstly, the time complexity of the proposed algorithm can be further reduced for faster computational speed in large-scale problems. The current PSO-SAVL, if applied in the optimization problems of more than 1000 dimensions, computational speed is slow. From this aspect, the modified ESE strategy proposed in ADDE may be a good choice to replace the original ESE strategy because it only utilizes the information of the globally best particle and the median best particle rather than complete positional information. Secondly, the uniform velocity limit for the whole population can be modified to realize different velocity limits of different particles for better performance. All particles in the current PSO-SAVL share the same velocity limit which may restrict further performance improvement.

# References

[1] R.C. Eberhart, J. Kennedy, A new optimizer using particle swarm theory, in: Proceedings of the Sixth International Symposium on Micro Machine and Human Science, IEEE, Nagoya, Japan. (1995) 39–43.

[2] R.-L. Tang, Y.-J. Fang, Modification of particle swarm optimization with human simulated property, Neurocomputing. 153 (2015) 319–331.

[3] R. Urraca, E. Sodupe-Ortega, J. Antonanzas, F. Antonanzas-Torres, F.J. Martinez-de-Pison, Evaluation of a novel GA-based methodology for model structure selection: The GA-PARSIMONY, Neurocomputing. 271 (2018) 9–17.

[4] Y. Wu, M. Gong, W. Ma, S. Wang, High-order graph matching based on ant colony optimization, Neurocomputing. 328 (2019) 97–104.

[5] Q. Wang, S. Chen, X. Luo, An adaptive latent factor model via particle swarm optimization, Neurocomputing. 369 (2019) 176–184.

[6] F. Hafiz, A. Swain, E.M.A.M. Mendes, Two-Dimensional (2D) particle swarms for structure selection of nonlinear systems, Neurocomputing. 367 (2019) 114–129.

[7] Y. Li, J. Xiao, Y. Chen, L. Jiao, Evolving deep convolutional neural networks by quantum behaved particle swarm optimization with binary encoding for image classification, Neurocomputing. 362 (2019) 156–165.

[8] F. Han, M.-R. Zhao, J.-M. Zhang, Q.-H. Ling, An improved incremental constructive single-hidden-layer feedforward networks for extreme learning machine based on particle swarm optimization, Neurocomputing. 228 (2017) 133–142.

[9] S. Goudarzi, W.H. Hassan, M.H. Anisi, A. Soleymani, M. Sookhak, M.K. Khan, A.-H.A. Hashim, M. Zareei, ABC-PSO for vertical handover in heterogeneous wireless networks, Neurocomputing. 256 (2017) 63–81.

[10] H. Ding, X. Gu, Hybrid of human learning optimization algorithm and particle swarm optimization algorithm with scheduling strategies for the flexible job-shop scheduling problem, Neurocomputing. 414 (2020) 313–332.

[11] B. Nasiri, M. Meybodi, M. Ebadzadeh, History-driven particle swarm optimization in dynamic and uncertain environments, Neurocomputing. 172 (2016) 356–370.

[12] B. Wang, S. Li, J. Guo, Q. Chen, Car-like mobile robot path planning in rough terrain using multi-objective particle swarm optimization algorithm, Neurocomputing. 282 (2018) 42–51.

[13] N. Zeng, Z. Wang, H. Zhang, K.-E. Kim, Y. Li, X. Liu, An improved particle filter with a novel hybrid proposal distribution for quantitative analysis of gold immunochromatographic strips, IEEE Trans. Nanotechnology. 18 (2019) 819–829.

[14] A. Ratnaweera, S.K. Halgamuge, H.C. Watson, Self-organizing hierarchical particle swarm optimizer with time-varying acceleration coefficients, IEEE Trans. Evol. Computat. 8 (2004) 240–255.

[15] Z.-H. Zhan, J. Zhang, Y. Li, H.S.-H. Chung, Adaptive particle swarm optimization, IEEE Trans. Syst., Man, Cybern. B. 39 (2009) 1362–1381.

[16] Z.-J. Wang, Z.-H. Zhan, S. Kwong, H. Jin, J. Zhang, Adaptive granularity learning distributed particle swarm optimization for large-scale optimization, IEEE Trans. Cybern. 51 (2021) 1175–1181.

[17] X. Xia, L. Gui, Z.-H. Zhan, A multi-swarm particle swarm optimization algorithm based on dynamical

topology and purposeful detecting, Applied Soft Computing. 67 (2018) 126-140.

[18] X. Xia, L. Gui, Y. Zhang, X. Xu, F. Yu, H. Wu, B. Wei, G. He, Y. Li, K. Li, A fitness-based adaptive differential evolution algorithm, Information Sciences. 549 (2021) 116–141.

[19] J.J. Liang, A.K. Qin, P.N. Suganthan, S. Baskar, Comprehensive learning particle swarm optimizer for global optimization of multimodal functions, IEEE Trans. Evol. Computat. 10 (2006) 281–295.

[20] W. Liu, Z. Wang, X. Liu, N. Zeng, D. Bell, A novel particle swarm optimization approach for patient clustering from emergency departments, IEEE Trans. Evol. Computat. 23 (2019) 632–644.

[21] X. Xia, L. Gui, F. Yu, H. Wu, B. Wei, Y.-L. Zhang, Z.-H. Zhan, Triple archives particle swarm optimization, IEEE Trans. Cybern. 50 (2020) 4862-4875.

[22] J.-Y. Li, Z.-H. Zhan, R.-D. Liu, C. Wang, S. Kwong, J. Zhang, Generation-level parallelism for evolutionary computation: a pipeline-based parallel particle swarm optimization, IEEE Trans. Cybern. (2020) 1–12.

[23] Z.-H. Zhan, J. Zhang, Y. Li, Y.-H. Shi, Orthogonal learning particle swarm optimization, IEEE Trans. Evol. Computat. 15 (2011) 832–847.

[24] L. Ma, S. Cheng, Y. Shi, Enhancing learning efficiency of brain storm optimization via orthogonal learning design, IEEE Trans. Syst. Man Cybern, Syst. (2020) 1–20.

[25] T. Huang, J. Huang, J. Zhang, An orthogonal local search genetic algorithm for the design and optimization of power electronic circuits, in: 2008 IEEE Congress on Evolutionary Computation, IEEE, Hong Kong, China. (2008) 2452–2459.

[26] W. Chu, X. Gao, S. Sorooshian, Handling boundary constraints for particle swarm optimization in high-dimensional search space, Information Sciences. 181 (2011) 4569–4581.

[27] S. Helwig, J. Branke, S. Mostaghim, Experimental analysis of bound handling techniques in particle swarm optimization, IEEE Trans. Evol. Computat. 17 (2013) 259–271.

[28] Y. Yan, R. Zhang, J. Wang, J. Li, Modified PSO algorithms with "Request and Reset" for leak source localization using multiple robots, Neurocomputing. 292 (2018) 82–90.

[29] H. Melo, J. Watada, Gaussian-PSO with fuzzy reasoning based on structural learning for training a Neural Network, Neurocomputing. 172 (2016) 405–412.

[30] X. Zhang, K.-J. Du, Z.-H. Zhan, S. Kwong, T.-L. Gu, J. Zhang, Cooperative coevolutionary bare-bones particle swarm optimization with function independent decomposition for large-scale supply chain network design with uncertainties, IEEE Trans. Cybern. 50 (2020) 4454–4468.

[31] X.-F. Liu, Z.-H. Zhan, Y. Gao, J. Zhang, S. Kwong, J. Zhang, Coevolutionary particle swarm optimization with bottleneck objective learning strategy for many-objective optimization, IEEE Trans. Evol. Computat. 23 (2019) 587–602.

[32] Z.-J. Wang, Z.-H. Zhan, W.-J. Yu, Y. Lin, J. Zhang, T.-L. Gu, J. Zhang, Dynamic group learning distributed particle swarm optimization for large-scale optimization and its application in cloud workflow scheduling, IEEE Trans. Cybern. 50 (2020) 2715–2729.

[33] Y. Shi, R.C. Eberhart, Parameter selection in particle swarm optimization, in: Proceedings of the 1998 International Conference on Evolutionary Programming VII. (1998) 591–600.

[34] S. Jiang, J. Shi, Y. Zhang, H. Han, Automatic test data generation based on reduced adaptive particle swarm

optimization algorithm, Neurocomputing. 158 (2015) 109–116.

[35] Y. Liu, Z. Qin, Z. Shi, J. Lu, Center particle swarm optimization, Neurocomputing. 70 (2007) 672–679.

[36] G. Wang, Variable velocity limit chaotic particle swarm optimization, in: The 2010 IEEE International Conference on Information and Automation, IEEE, Harbin, China. (2010) 1661–1666.

[37] J. Barrera, O. Álvarez Bajo, J.J. Flores, C.A. Coello Coello, Limiting the velocity in the particle swarm optimization algorithm, CyS. 20 (2016).

[38] A.O. Adewumi, A.M. Arasomwan, Improved particle swarm optimizer with dynamically adjusted search space and velocity limits for global optimization, Int. J. Artif. Intell. Tools. 24 (2015) 1550017.

[39] M. Pluhacek, R. Senkerik, A. Viktorin, T. Kadavy, Study on velocity clamping In PSO using CEC'13 benchmark, in: ECMS 2018 Proceedings Edited by Lars Nolle, Alexandra Burger, Christoph Tholen, Jens Werner, Jens Wellhausen, ECMS. (2018) 150–155.

[40] M. Yusoff, J. Ariffin, A. Mohamed, DPSO based on a min-max approach and clamping strategy for the evacuation vehicle assignment problem, Neurocomputing. 148 (2015) 30–38.

[41] R. Kundu, S. Das, R. Mukherjee, S. Debchoudhury, An improved particle swarm optimizer with difference mean based perturbation, Neurocomputing. 129 (2014) 315–333.

[42] M. Clerc, J. Kennedy, The particle swarm - explosion, stability, and convergence in a multidimensional complex space, IEEE Trans. Evol. Computat. 6 (2002) 58–73.

[43] Z.-H. Zhan, Z.-J. Wang, H. Jin, J. Zhang, Adaptive distributed differential evolution, IEEE Trans. Cybern. 50 (2020) 4633–4647.

[44] Y. Shi, R.C. Eberhart, Empirical study of particle swarm optimization, in: Proceedings of the 1999 Congress on Evolutionary Computation-CEC99 (Cat. No. 99TH8406), IEEE, Washington, DC, USA. (1999) 1945–1950.

[45] J.J. Liang, P.N. Suganthan, K. Deb, Novel composition test functions for numerical global optimization, in: Proceedings 2005 IEEE Swarm Intelligence Symposium, 2005. SIS 2005., IEEE, Pasadena, CA, USA. (2005) 68–75.

[46] T. Peram, K. Veeramachaneni, C.K. Mohan, Fitness-distance-ratio based particle swarm optimization, in: Proceedings of the 2003 IEEE Swarm Intelligence Symposium. SIS'03 (Cat. No.03EX706), IEEE, Indianapolis, IN, USA. (2003) 174–181.

[47] K.E. Parsopoulos, M.N. Vrahatis, Unified particle swarm optimization in dynamic environments, in: Applications of Evolutionary Computing, Springer Berlin Heidelberg, Berlin, Heidelberg. (2005) 590–599.

[48] J. Sun, B. Feng, W. Xu, Particle swarm optimization with particles having quantum behavior, in: Proceedings of the 2004 Congress on Evolutionary Computation (IEEE Cat. No.04TH8753), IEEE, Portland, OR, USA. (2004) 325–331.

[49] Y. Wang, B. Li, T. Weise, J. Wang, B. Yuan, Q. Tian, Self-adaptive learning based particle swarm optimization, Information Sciences. 181 (2011) 4515–4538.

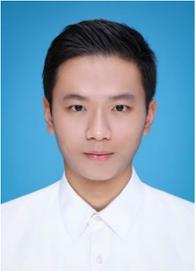
**Dr. Xinze Li** received his bachelor's degree in Electrical Engineering and its Automation from Shandong University, China, 2018. He received his Ph.D. degree in Electrical and Electronic Engineering from Nanyang Technological University, Singapore, 2023.

His research interests include dc-dc converter, modulation design, digital twins for power electronics systems, design process automation, light and explainable AI for power electronics with physics-informed systems, application of AI in power electronics, and deep learning and machine learning algorithms.

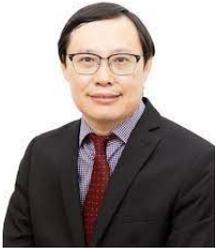
**Dr. Kezhi Mao** obtained his BEng, MEng and PhD from University of Jinan, Northeastern University, and University of Sheffield in 1989, 1992 and 1998 respectively. Since then, he has been working at School of Electrical and Electronic Engineering, Nanyang Technological University, Singapore, and is now an Associate Professor. His research covers a couple of subfields of artificial intelligence (AI), including machine learning, image processing, natural language processing and information fusion. Over the past 20 years, he has developed novel algorithms and frameworks to address various problems in AI. As a strong advocate of translational research, he has collaborated with government agencies and hospitals and developed a couple of prototypes of AI systems for image processing and natural language processing. He now serves as Member of Editorial Board of Neural Networks, Academic Editor of Computational Intelligence and Neuroscience, and General Chair and General Co-Chair of a number of international conferences.

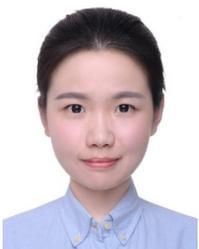
**Dr. Fanfan Lin** Fanfan Lin was born in Fujian, China in 1996. She received her bachelor degree in electrical engineering from Harbin Institute of Technology in China in 2018. She has been awarded the Joint Ph. D. degree in Nanyang Technological University, Singapore and Technical University of Denmark, Denmark, in 2023. Her research interest includes large language models for the design of large-scale power electronics systems, multi-modal AI for the maintenance of power converters, and the application of AI in power electronics.

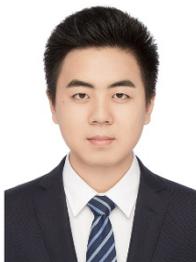
**Zijian Feng** received a B.Sc. degree from School of Electronics and Information Technology, Sun Yat-Sen University, Guangzhou, China, in 2019, and an M.Sc. degree in Signal Processing from School of Electrical and Electronic Engineering, Nanyang Technological University, Singapore, in 2020. He is currently pursuing a Ph.D. degree with the Interdisciplinary Graduate School, Nanyang Technological University, Singapore. His research interests include machine learning and natural language processing.